\pdfoutput=1
\documentclass[10pt,twocolumn,letterpaper]{article}

\usepackage[dvipsnames,table]{xcolor}
\usepackage[pagenumbers]{cvpr}

\usepackage[T1]{fontenc}
\usepackage[utf8]{inputenc}
\usepackage{microtype}
\usepackage{multirow}
\usepackage{mathtools}
\usepackage{enumitem}
\usepackage{array}
\usepackage{siunitx}
\usepackage{algorithm}
\usepackage{algpseudocode}
\usepackage[most]{tcolorbox}
\algrenewcommand\algorithmicrequire{\textbf{Input:}}
\usepackage{amsmath,amssymb,graphicx,booktabs,array,tabularx,multirow}
\usepackage[scaled]{helvet}
\usepackage{xcolor,tikz,float}\usetikzlibrary{arrows.meta,positioning,fit,calc,backgrounds}
\definecolor{mblue}{HTML}{BDD7EE}\definecolor{mblueL}{HTML}{2E75B6}
\definecolor{mpink}{HTML}{F8CBAD}\definecolor{mpinkL}{HTML}{ED7D31}
\definecolor{mgreen}{HTML}{C6E0B4}\definecolor{mgreenL}{HTML}{548235}
\definecolor{myell}{HTML}{FFE699}\definecolor{myellL}{HTML}{BF9000}
\definecolor{mgray}{HTML}{F2F2F2}\definecolor{mgrayL}{HTML}{7F7F7F}
\definecolor{mred}{HTML}{F4B8B8}\definecolor{mredL}{HTML}{C00000}\definecolor{ink}{HTML}{3B3B3B}
\tikzset{mini/.style={rounded corners=4pt,line width=.55pt,inner xsep=2mm,inner ysep=1.4mm,align=center,font=\sffamily\scriptsize},enc/.style={mini,fill=mblue,draw=mblueL},tgt/.style={mini,fill=mpink,draw=mpinkL},prd/.style={mini,fill=mgreen,draw=mgreenL},act/.style={mini,fill=myell,draw=myellL},lat/.style={mini,fill=mgray,draw=mgrayL},ar/.style={-{Latex[length=1.8mm]},line width=.6pt,draw=ink},ema/.style={-{Latex[length=1.8mm]},line width=.55pt,draw=mpinkL,dashed}}
\newcommand{\pill}[2]{\tikz[baseline=-0.65ex]\node[rounded corners=2pt,fill=#1!16,draw=#1!65!black,line width=.35pt,inner xsep=2.2pt,inner ysep=1.2pt,font=\sffamily\fontsize{6.7}{7.2}\selectfont]{#2};}

\definecolor{cvprblue}{rgb}{0.21,0.49,0.74}
\newtcolorbox{algobox}[2][]{enhanced,breakable,colback=white,colframe=black!60,
  coltitle=white,colbacktitle=black!60,fonttitle=\bfseries,title={#2},
  boxrule=0.5pt,arc=2pt,left=4pt,right=4pt,top=3pt,bottom=3pt,#1}
\newcommand{\algcomment}[1]{\Statex{\footnotesize\hspace{1em}$\triangleright$~#1}}

\usepackage[pagebackref,breaklinks,colorlinks]{hyperref}
\hypersetup{pdftitle={World Models for Cross-Machine CNC Transfer under Partial Sensor Overlap},pdfauthor={Ayoub Louaye Bouaziz, Matthieu Ostertag, Anton Demasles},pdfsubject={arXiv preprint, cs.LG},pdfkeywords={world models, command-conditioned prediction, CNC, cross-machine transfer, schema shift, instance normalization, JEPA}}

\title{World Models for Cross-Machine CNC Transfer under Partial Sensor Overlap}
\author{Ayoub Louaye Bouaziz$^{1}$ \qquad Matthieu Ostertag$^{2}$ \qquad Anton Demasles$^{2}$\\[3pt]
{\normalsize $^{1}$Universit\'e de Bretagne Occidentale, Brest, France \qquad $^{2}$Mines Nancy, Universit\'e de Lorraine, Nancy, France}\\[3pt]
{\tt\footnotesize\hspace*{-\tabcolsep}abouaziz@univ-brest.fr \; matthieu.ostertag7@etu.univ-lorraine.fr \; anton.demasles2@etu.univ-lorraine.fr\hspace*{-\tabcolsep}}}
\date{}
\begin{document}
\maketitle

\begin{abstract}
Industrial world models must move between machines whose dynamics, sensing interfaces and command conventions differ. This study asks whether a command-conditioned latent world model---trained to predict future representations of the process rather than to reconstruct future samples---keeps its value on a machine it has never seen: a source CNC machine exposes 17 sensor channels, the target sharing 10 of those. All model selection uses source data only, and the locked configuration is evaluated on the target once. Two findings follow. First, latent-predictive pretraining brings no in-domain forecasting gain over matched training from scratch, so source accuracy alone cannot show what such a representation is worth. Second, the transferred model beats persistence on the unseen machine (with $R^2\approx0.01$ against the target mean) but trails official forecasters that normalize each input window by its own statistics; a post-lock ablation, declared before it ran, shows that this input normalization alone closes the gap, and closing it costs predictive calibration. Cross-machine transfer under partial sensor overlap is therefore a distinct evaluation axis for command-conditioned world models.
\end{abstract}

\section{Introduction}
\begin{figure*}[t]
  \centering
  \includegraphics[width=\textwidth]{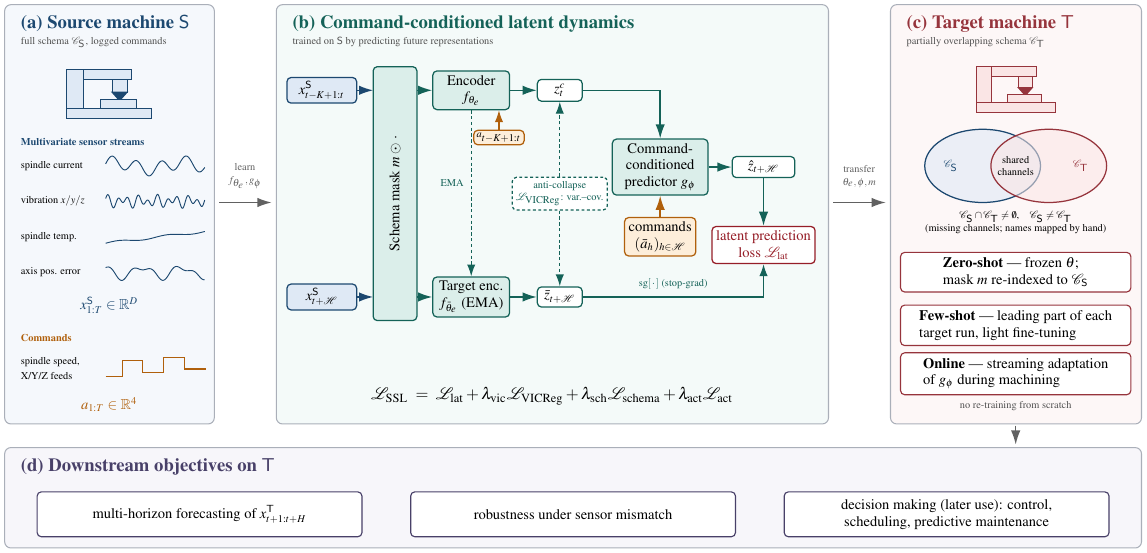}
  \caption{\textbf{Overview of the command-conditioned world model for cross-machine CNC transfer.}
  The model is joint-embedding predictive: it predicts the embedding of a future window rather than its samples; command-conditioned: each prediction is made under a given sequence of future commands; and schema-adaptive: it reads whichever channels a machine provides, marked by a presence mask, with the same weights. A source CNC machine $\mathsf S$ provides multivariate sensor streams and commands, whereas the target machine $\mathsf T$ exposes only a partially overlapping sensor schema. The model learns command-conditioned latent dynamics using a context encoder, an EMA target encoder, latent prediction, schema masking, and anti-collapse regularization. The learned representation is evaluated zero-shot and after few-shot adaptation on the target machine (the few-shot results come from the pre-lock model); the online setting of panel~(c) is described but not evaluated in this paper. The intended downstream use is cross-machine forecasting under machine and sensor-schema shift, with decision support as a later application.}
  \label{fig:overview}
\end{figure*}

Industrial control systems generate multivariate sensor streams while receiving commands, i.e.\ the spindle-speed and feed setpoints issued by the numerical controller. A world model of such a process predicts how the measured signals will evolve under a given sequence of future commands. To be deployable, it must keep working when it moves to another machine, which can differ from the training machine in two ways. Under \emph{domain shift}, the list of measured channels is the same but the signals differ: dynamics, levels, units and sampling regimes change. Under \emph{schema shift}, the list of channels itself changes, so part of the input that a trained model expects does not exist on the new machine. A CNC target machine typically exhibits both at once, which separates this problem from in-domain time-series forecasting.

Joint-Embedding Predictive Architectures (JEPAs) \citep{assran2023ijepa,bardes2024vjepa} are trained to predict, from a context, the embedding that an encoder assigns to a future or masked part of the input, instead of reconstructing that part sample by sample. The predicted embedding must carry what is needed to anticipate the process---here the operating regime, the load and the position in the cutting cycle---and may discard sensor-specific detail, such as noise, offsets and the fine waveform of each channel, that a reconstruction loss would have to reproduce. That detail is precisely what does not carry over to another machine. Our working hypothesis is therefore that such a representation remains useful when the model moves to a machine it was not trained on. The hypothesis has support in other applications: feature prediction yields representations that transfer well across image and video tasks \citep{assran2023ijepa,bardes2024vjepa}, and a JEPA latent dynamics model used for planning generalises to held-out maze layouts better than model-free alternatives \citep{sobal2025pldm}. There, however, the agent's physics and observation space are unchanged and only the wall layout varies; here both the dynamics and the sensor schema change. This paper puts the hypothesis under a sealed cross-machine test, but it does not isolate the latent objective on the target, since every evaluated arm is trained with it (Sec.~\ref{sec:limitations}). The test must also contend with known difficulties: JEPA training is prone to representational collapse \citep{bardes2022vicreg}, strong forecasting and masked-pretraining baselines already exist \citep{dong2023simmtm,nie2023patchtst,liu2024itransformer}, and action-aware world models have been studied for industrial process control \citep{yan2025actionable}.

We test this hypothesis on a CNC source-to-target pair (Figure~\ref{fig:overview}) built from two public datasets: the THWS five-axis CNC milling production dataset with multiple changeovers \citep{martinez2025thws} as the source machine, and the FH JOANNEUM CNC machining data repository with high-frequency energy measurements \citep{brillinger2025joanneum} as the target machine. The source machine records 17 canonical sensor channels across 62 program sessions; the target shares 10 of those channels across 7 independent runs. The test follows the cross-machine evaluation protocol of Sec.~\ref{sec:protocol}: all model selection uses source data only, after which the locked model is evaluated on the sealed target machine in a single declared pass (an earlier, pre-lock model was also evaluated on the target, Sec.~\ref{sec:fewshot}, and a post-lock ablation, declared before it ran, reads it again, Sec.~\ref{sec:revin-results}). The protocol also includes source-only normalization, session-grouped splitting, held-out SSL validation, unit audits, collapse diagnostics, matched scratch/pretraining controls, trivial forecasting baselines, and run-level target statistics.

The scope is deliberately narrow. The paper studies a command-conditioned latent dynamics model and evaluates it by multi-horizon forecasting under given command sequences, with a diagnostic that checks that the forecasts depend on the commands. It learns no policy, does no planning, and closes no control loop.

\paragraph{Contributions.} The contributions are as follows:
\begin{itemize}[leftmargin=*,nosep]
    \item a leakage-audited cross-machine CNC protocol combining machine shift with partial sensor overlap;
    \item a schema-adaptive, command-conditioned JEPA world model: every machine is read through one canonical channel list with a presence mask, and attention and pooling run over present channels only, so one set of weights reads the 17-channel source and the 10-channel target without retraining, reshaping or imputing; it is trained with latent schema consistency, EMA targets and explicit anti-collapse regularization;
    \item an empirical diagnosis showing that JEPA pretraining does not improve clean source RMSE over scratch, while representation stability is sensitive to latent-loss weighting, variance--covariance regularization, and EMA dynamics;
    \item a pre-lock diagnostic adaptation sweep suggesting that limited target support improves cross-machine forecasting relative to zero-shot transfer; it is retained as diagnostic rather than confirmatory evidence;
    \item a post-lock paired ablation of input scaling: the official baselines rescale each input window by its own mean and standard deviation and undo the rescaling on the output (RevIN \citep{kim2022revin}); adding the same step to our model, computed over present channels only, closes the zero-shot RMSE gap ($0.495\pm0.004$ against 0.503 and 0.498) but breaks its uncertainty estimates on near-stationary windows (target NLL 20.6), so input scaling is sufficient to explain the gap.
\end{itemize}

\section{Related Work}
\paragraph{Joint-embedding predictive learning.}
I-JEPA introduced representation prediction as a non-generative self-supervised objective for images \citep{assran2023ijepa}, later extended to video \citep{bardes2024vjepa}; VICReg regularizes representation variance and covariance to avoid collapse \citep{bardes2022vicreg}. The CNC experiments below show that anti-collapse regularization is operationally necessary rather than auxiliary.

\paragraph{JEPA world models.}
PLDM trains a JEPA latent dynamics model on reward-free offline data and plans with it by MPPI trajectory optimization; it prevents collapse with VICReg-style variance and covariance terms, a temporal-smoothness term and a one-step inverse-dynamics loss that regresses $a_t$ from $(z_t,z_{t+1})$, and it generalises best to held-out maze layouts, where the agent's point-mass physics and observation space are unchanged and only the wall layout varies \citep{sobal2025pldm}; an unchanged observation space is exactly what schema shift breaks. The EB-JEPA library builds action-conditioned JEPA world models with the same family of regularizers; in its randomized-wall Two Rooms ablation, removing the inverse-dynamics loss collapses the representation and planning success falls from 97\% to 1\% \citep{terver2026ebjepa}. In both works the inverse-dynamics model is a training regularizer and is not used at planning time. LeWorldModel removes this machinery: it trains end-to-end from pixels with only a next-embedding prediction loss and a Gaussian-embedding regularizer (SIGReg), without inverse dynamics, EMA or stop-gradient, which leaves one loss weight to tune instead of PLDM's six \citep{maes2026lewm}. Our command-recovery head $r_\omega$ (Sec.~\ref{sec:predictor}) belongs to the inverse-dynamics family but differs in four ways: it is multi-step, regressing the mean command over horizon $h$ from the context latent and the EMA-target latent; it is lightly weighted ($\lambda_{\mathrm{act}}=0.05$); because the target latent is detached, it shapes only the context branch; and, unlike PLDM and EB-JEPA, whose encoders see observations only, our context latent is computed from the past commands as well, so the term is a weaker constraint than a one-step inverse-dynamics loss. It sits alongside a variance--covariance penalty and a schema-consistency term, and Sec.~\ref{sec:hparams} shows that the stability of such multi-term objectives depends on the loss weighting, the difficulty LeWorldModel addresses.

\paragraph{Time-series forecasting and masked pretraining.}
SimMTM reconstructs masked series from complementary masked neighbors \citep{dong2023simmtm}; PatchTST uses channel-independent patch tokens \citep{nie2023patchtst}; iTransformer inverts the tokenization axis to model multivariate dependencies \citep{liu2024itransformer}; RevIN counters distribution shift by instance normalization \citep{kim2022revin}; Sec.~\ref{sec:revin} tests it inside our model under sensor presence masks. These methods motivate a protocol in which JEPA is not assumed to dominate source-domain forecasting and must justify its value under transfer and schema mismatch.

\paragraph{World models and industrial adaptation.}
Latent dynamics models support multi-step prediction and planning \citep{hafner2019planet}, and actionable world models have been studied for industrial process control \citep{yan2025actionable}. The present setting adds cross-machine transfer, partial sensor overlap, explicit command alignment, source-only normalization, and post-training adaptation on the target machine.

\section{Problem Formulation}
Two CNC machines are involved: a source machine $\mathsf S$, on which all training and model selection take place, and a target machine $\mathsf T$, used only for transfer. Their channel sets $\mathcal C_{\mathsf S}$ and $\mathcal C_{\mathsf T}$ overlap only partially: $|\mathcal C_{\mathsf S}|=17$, and $\mathsf T$ shares 10 of these channels; target-only channels are discarded. On both machines the observation $x_t\in\mathbb R^{D}$, $D=17$, lists the channels of $\mathcal C_{\mathsf S}$ in one fixed order; a superscript, $x_t^{\mathsf S}$ or $x_t^{\mathsf T}$, is added only where the machine matters. The presence mask $m_t\in\{0,1\}^{D}$ marks the entries actually measured: on $\mathsf T$ the seven channels of $\mathcal C_{\mathsf S}\setminus\mathcal C_{\mathsf T}$ always have $m=0$, and training-time masking (Sec.~\ref{sec:encoder}) sets further entries of $m$ to $0$. The command $a_t\in\mathbb R^4$ holds the spindle-speed and X/Y/Z feed setpoints issued by the numerical controller and logged at step $t$; it is taken to act over the interval $(t-1,t]$ that ends at $t$, so $a_{t+1}$ is the first command that acts on the future rows. The world-model literature calls such inputs actions; we call them commands, the setpoints a numerical controller issues to its drives.

Given the $K$ rows ending at $t$ and the commands up to $t+H$, where $H$ is the longest horizon, we want to learn the conditional probability
\begin{equation}\label{eq:target}
    p_\theta\!\left(x_{t+1:t+H}\mid x_{t-K+1:t},m_{t-K+1:t},a_{t-K+1:t+H}\right)
\end{equation}
as represented by a neural network with parameters $\theta$. Eq.~\eqref{eq:target} is the object of interest, not the training loss: Sec.~\ref{sec:method} factors it through a latent state, and Sec.~\ref{sec:head} writes the resulting model explicitly.

The mask separates an absent channel from a channel that reads zero. Both are stored as the value $0$, but the absent channel carries $m=0$ and the zero reading $m=1$. The encoder builds one token per channel and time step and drops every token with $m=0$: such a token takes no part in the cross-channel attention and no part in the average that reduces a time step to one vector (when a step has no present channel, a single channel is kept so that the average is defined). An absent channel therefore contributes nothing rather than a zero, and on $\mathsf T$ the seven source-only channels simply do not exist for the model, which runs with the same weights on both machines.

\begin{figure*}[t]
  \centering
  \includegraphics[width=\textwidth]{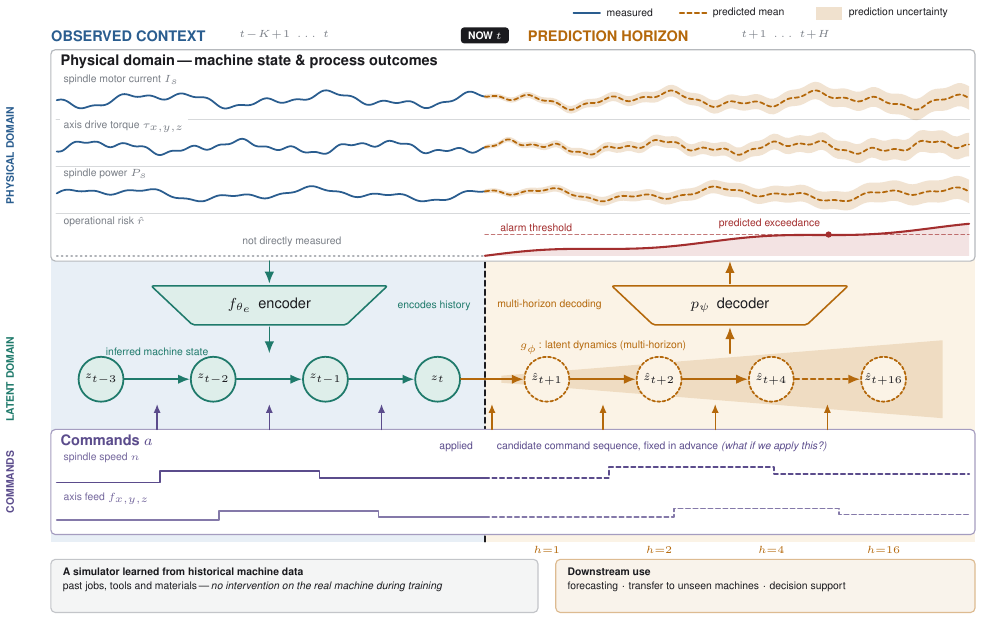}
  \caption{\textbf{A CNC world model in practice.} The schematic illustrates the operational interpretation of the learned dynamics model. Recent sensor measurements are encoded into a latent state, future commands condition multi-horizon latent predictions, and a physical decoder maps predicted latent states onto multi-horizon process variables and uncertainty. The candidate command sequence is given in advance: it is an input to the model, not an output. Nothing here is closed loop---no prediction is fed back to the machine, and no controller is closed around the model anywhere in this paper. Its purpose is to distinguish passive forecasting from command-conditioned simulation of candidate futures. The chain of latent states and the operational-risk lane are illustrative: the implementation predicts all horizons in one pass from the context latent (Sec.~\ref{sec:predictor}), and the risk variable is a downstream use, not an output of the model in this paper.}
  \label{fig:worldmodel}
\end{figure*}

The forecasting metric measures numerical prediction error. The world-model requirement is stronger: predictions should also respond meaningfully to changes in future commands. This distinction is essential in highly autocorrelated industrial streams, where a model can obtain low short-horizon error while effectively ignoring the commands. Figure~\ref{fig:worldmodel} illustrates this operational interpretation. The figure is read as three lanes sharing one time axis that is split at \emph{now}: a physical lane carrying the measured and the predicted process variables together with an illustrative alarm threshold, a latent lane carrying the encoder, the latent dynamics and the decoder, and a command lane carrying the already applied and the candidate command sequences.

\section{Method}\label{sec:method}
\subsection{Schema-adaptive context encoding}\label{sec:encoder}
The context encoder $f_{\theta_e}$ receives the observed values, the presence mask and the past commands:
\begin{equation}\label{eq:enc}
    z_t^c=f_{\theta_e}(x_{t-K+1:t},m_{t-K+1:t},a_{t-K+1:t}).
\end{equation}
Schema masking generates multiple partial views of the same source trajectory. Missing channels are represented structurally rather than replaced before normalization, so the model receives an explicit indication of which measurements are available.

The encoder $f_{\theta_e}$ works in two passes. The first acts within each time step: a transformer attends over the tokens of the channels present at that step, a mean over the same tokens reduces them to one vector, and a linear map of that step's command is added to it. Because this pass operates on a set of present-channel tokens of any size, the same weights serve any subset of sensors, which is what lets one encoder read both machines. The second pass is a temporal transformer over the $K$ step vectors and carries the dynamics; $z^c_t$ is its output at the last position. The locked configuration uses $d=256$, eight attention heads, six temporal layers and pre-norm blocks (\hyperref[app:config]{App.~G}).

\subsection{Command-conditioned latent prediction}\label{sec:predictor}
A predictor $g_\phi$ receives the context representation and candidate future commands:
\begin{equation}\label{eq:pred}
    \hat z_{t+\mathcal H}=g_\phi\big(z_t^c,(\bar a_h)_{h\in\mathcal H}\big),
\end{equation}
where $\hat z_{t+\mathcal H}=(\hat z_{t+h})_{h\in\mathcal H}$. In practice the context spans $K=32$ steps and prediction is made at a fixed set of horizons $\mathcal{H}=\{1,2,4,8,16\}$, $H=16$, by direct multi-horizon prediction rather than autoregressive rollout. Windows are formed on the 1\,Hz grid (Table~\ref{tab:data}), so the context covers 32\,s and the longest horizon 16\,s. For each $h\in\mathcal{H}$ the predictor receives $\bar a_h=\mathrm{mean}(a_{t+1:t+h})$, the mean command over $(t,t+h]$; because these means are nested over 1, 2, 4, 8 and 16 steps, together they encode the level of the future command and a coarse profile of it, not its step-by-step sequence. All horizon slots are filled in a single pass from the same $z^c_t$: every slot receives the context representation, a horizon embedding and its own mean command, and a transformer runs over the slots with a causal mask, so $\hat z_{t+h}$ depends on $z^c_t$ and on $\bar a_{h'}$ for $h'\le h$ only; no predicted latent is fed back as an input. The slots are the horizons of the evaluation grid, so training and evaluation share the same prediction structure. The factorization over horizons of the physical model, Eq.~\eqref{eq:chain}, is a modelling choice of the head (Sec.~\ref{sec:head}), not a consequence of this structure.
A target encoder $f_{\bar\theta_e}$, with the architecture of $f_{\theta_e}$, embeds the future rows $x_{t+\mathcal H}=\{x_{t+h}\}_{h\in\mathcal H}$ into $\bar z_{t+h}$; the five target rows are encoded jointly, with zero commands, so $\bar z_{t+h}$ depends on all of them, up to $x_{t+16}$. Its parameters are updated by exponential moving average,
\begin{equation}\label{eq:ema}
    \bar\theta_e \leftarrow \tau\bar\theta_e+(1-\tau)\theta_e,
\end{equation}
and gradients are stopped through the target branch, written $\mathrm{sg}[\cdot]$.

The SSL objective is
\begin{equation}\label{eq:lssl}
\mathcal{L}_{\mathrm{SSL}}=
\lambda_{\mathrm{lat}}\mathcal{L}_{\mathrm{lat}}+
\lambda_{\mathrm{vic}}\mathcal{L}_{\mathrm{VICReg}}+
\lambda_{\mathrm{sch}}\mathcal{L}_{\mathrm{schema}}+
\lambda_{\mathrm{act}}\mathcal{L}_{\mathrm{act}},
\end{equation}
with, averaged over the mini-batch,
\begin{align}
\mathcal L_{\mathrm{lat}}&=\tfrac{1}{|\mathcal H|}\textstyle\sum_{h}\ell\big(\hat z_{t+h},\mathrm{sg}[\bar z_{t+h}]\big),\label{eq:llat}\\
\mathcal L_{\mathrm{VICReg}}&=\tfrac12\big[V(z^c_{\mathrm{pool}})+\tfrac{1}{|\mathcal H|}\textstyle\sum_{h}V(\mathrm{sg}[\bar z_{t+h}])\big],\label{eq:lvic}\\
\mathcal L_{\mathrm{schema}}&=\tfrac{1}{|\mathcal H|}\textstyle\sum_{h}\ell\big(\hat z'_{t+h},\mathrm{sg}[\hat z_{t+h}]\big),\label{eq:lsch}\\
\mathcal L_{\mathrm{act}}&=\tfrac{1}{4|\mathcal H|}\textstyle\sum_{h}\big\|r_\omega(z^c_t,\mathrm{sg}[\bar z_{t+h}])-\bar a_h\big\|_2^2,\label{eq:lact}
\end{align}
where sums run over $h\in\mathcal H$, $\ell$ is the smooth-$L_1$ loss averaged over the $d$ latent coordinates, $z^c_{\mathrm{pool}}$ is the mean of the $K$ outputs of the temporal transformer, $\hat z'_{t+h}$ is the prediction from the reduced view described below, and
\begin{equation}\label{eq:vic}
\begin{split}
V(Z)={}&\tfrac{\alpha}{d}\textstyle\sum_{j}\max\!\big(0,\,1-\sqrt{\mathrm{Var}(Z_{:j})+\varepsilon}\big)\\
&+\tfrac{\beta}{d}\textstyle\sum_{i\neq j}C_{ij}(Z)^2
\end{split}
\end{equation}
is computed over the batch of $d$-dimensional latents $Z$, with $C(Z)$ their covariance, $\varepsilon$ a small constant and $(\alpha,\beta)=(25,1)$ (\hyperref[app:config]{App.~G}): a variance hinge that penalizes a standard deviation below $1$ in any coordinate, and a covariance penalty that decorrelates the coordinates. VICReg's invariance term is omitted because $\mathcal L_{\mathrm{lat}}$ already aligns the two branches. Since the target branch is detached, only the context half of \eqref{eq:lvic} carries gradient; the target half is a monitored constant.

Only three of the four weights are free. In the pretraining stage a common rescaling of all four multiplies the gradient by a scalar, which AdamW's normalized update absorbs up to its $\epsilon$ and the gradient-norm clip; we therefore fix $\lambda_{\mathrm{lat}}=1$ as a normalization and read the other weights as ratios to it. In the second stage the common scale also sets the balance against the Gaussian negative log-likelihood, whose weight is $1$ (\hyperref[app:config]{App.~G}). The search (\hyperref[app:search]{App.~A}) and the $\lambda_{\mathrm{lat}}$ sweep of Sec.~\ref{sec:hparams} vary $\lambda_{\mathrm{lat}}$ with the other weights held fixed, so there $\lambda_{\mathrm{lat}}$ is the ratio of the latent term to the auxiliary terms; $\lambda_{\mathrm{lat}}=0$ removes the latent term altogether.

\begin{figure*}[t]
  \centering
  \includegraphics[width=0.72\textwidth]{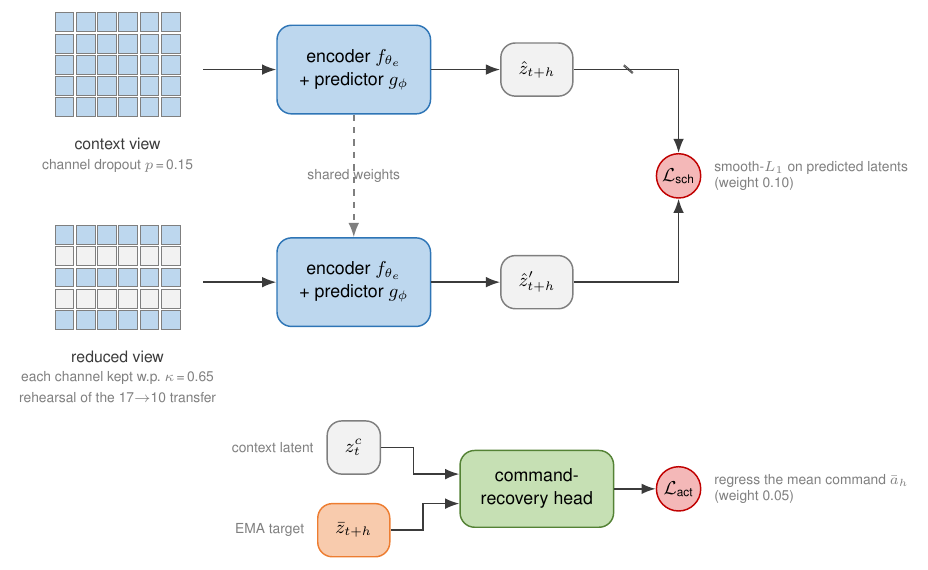}
  \caption{\textbf{The two auxiliary objectives of the locked candidate.} \emph{Schema consistency}: the same window is processed twice through the shared encoder and predictor, once with the channel-masked context view and once with a further random channel subset of it (each channel kept with probability $\kappa=0.65$, at least one channel kept); the predicted latents of the subset view are pulled toward the detached predicted latents of the full view with a smooth-$L_1$ penalty ($\lambda_{\mathrm{sch}}=0.10$), which exposes the model to reduced-schema views during source training. \emph{Command recovery}: a small head reads the context latent $z^c_t$ and the EMA-target latent $\bar z_{t+h}$ and regresses the mean command $\bar a_h$ applied over $(t,t+h]$ ($\lambda_{\mathrm{act}}=0.05$), which pressures the context latent to retain the command.}
  \label{fig:mechanisms}
\end{figure*}

Schema consistency is imposed on latent dynamics rather than on an untrained physical output head: during pretraining the head has no supervised target and receives no gradient, so a constraint on its output would be empty. In \eqref{eq:lsch}, $\hat z_{t+h}$ is the prediction from the channel-masked context view already used by $\mathcal L_{\mathrm{lat}}$, and $\hat z'_{t+h}$ the prediction from a further random subset of that view's present channels, each kept with probability $\kappa=0.65$, at least one kept. Combined with the channel dropout of the context view ($p_{\mathrm{drop}}=0.15$), a source channel is absent from the reduced view with probability $1-0.85\times0.65\approx0.45$, close to the fraction $7/17\approx0.41$ of source channels the target machine lacks. The reference $\hat z_{t+h}$ is detached, which makes the constraint one-sided: only the reduced view is pulled toward the fuller one, and the fuller view is never dragged toward the poorer prediction. In \eqref{eq:lact}, a small head $r_\omega$ infers the mean command $\bar a_h$ from the pair $(z^c_t,\bar z_{t+h})$; it plays the role of an inverse-dynamics model on latent states, in a weak form: the context encoder adds each step's command to its token and CNC commands are piecewise constant, so $\bar a_h$ is largely recoverable from $z^c_t$ alone. Because $\bar z_{t+h}$ comes from the detached target branch, the only trainable path runs through $z^c_t$: the term pressures $z^c_t$ to retain what identifies the applied command. Its effect is not isolated by an ablation; the matched candidate with schema consistency but without command recovery reaches a shuffled-command ratio of $1.044$ (four seeds, \hyperref[app:outcomes]{App.~C}) against $1.058$ for the locked candidate (seven seeds). The selection score of Sec.~\ref{sec:selection} carries the matching diagnostic: a candidate whose source-validation RMSE under shuffled future commands is less than $1.02$ times its RMSE under true commands is flagged as making weak use of the command. Figure~\ref{fig:mechanisms} shows both mechanisms.

\refstepcounter{algorithm}\label{alg:train}
\begin{algobox}{Algorithm \thealgorithm: self-supervised pretraining step (one mini-batch)}
\footnotesize
\begin{algorithmic}[1]
\Require windows of context rows $x_{t-K+1:t}$, masks $m_{t-K+1:t}$ and commands $a_{t-K+1:t}$, with targets $\{x_{t+h},\,a_{t+1:t+h}\}_{h\in\mathcal H}$; context encoder $f_{\theta_e}$; predictor $g_\phi$; command-recovery head $r_\omega$; EMA target encoder $f_{\bar\theta_e}$; channel-drop probability $p_{\mathrm{drop}}$; view keep-probability $\kappa$; weights $\lambda_{\mathrm{lat}},\lambda_{\mathrm{vic}},\lambda_{\mathrm{sch}},\lambda_{\mathrm{act}}$; momentum $\tau$
\State $\bar a_h \gets \mathrm{mean}(a_{t+1:t+h})$ for $h\in\mathcal H$
\algcomment{mean command over $(t,t+h]$; nested means give a coarse profile}
\State $(\tilde x,\tilde m) \gets \textsc{Corrupt}(x_{t-K+1:t}, m_{t-K+1:t};p_{\mathrm{drop}})$
\algcomment{each channel dropped for the whole window with probability $p_{\mathrm{drop}}$; dropped entries set to $0$ and marked absent}
\State \emph{RevIN variant only:} $(c,s)\gets\textsc{Stats}(\tilde x,\tilde m)$; $\tilde x\gets(\tilde x-c)/s$; $x_{t+\mathcal H}\gets(x_{t+\mathcal H}-c)/s$
\algcomment{per-channel context-window statistics over present entries; identity for unseen channels (Sec.~\ref{sec:revin})}
\State $z^c_t \gets f_{\theta_e}(\tilde x,\tilde m,a_{t-K+1:t})$
\algcomment{context latent}
\State $\hat z_{t+\mathcal H} \gets g_\phi\big(z^c_t,(\bar a_h)_{h\in\mathcal H}\big)$
\algcomment{one pass over the horizon slots, causal slot mask}
\State $\bar z_{t+\mathcal H} \gets \mathrm{sg}\big[f_{\bar\theta_e}(x_{t+\mathcal H}, m_{t+\mathcal H}, \mathbf 0)\big]$
\algcomment{the target rows are encoded jointly, with zero commands; stop-gradient}
\State $\mathcal L_{\mathrm{lat}}$ by \eqref{eq:llat}
\State $\mathcal L_{\mathrm{VICReg}}$ by \eqref{eq:lvic} and \eqref{eq:vic}
\algcomment{only the context half carries gradient}
\State $\mathcal L_{\mathrm{act}}$ by \eqref{eq:lact}
\algcomment{command recovery; the trainable path runs through $z^c_t$}
\State $(\tilde x',\tilde m') \gets \textsc{SchemaView}(\tilde x,\tilde m;\kappa)$
\algcomment{keep each channel with probability $\kappa$, at least one}
\State $\hat z'_{t+\mathcal H} \gets g_\phi\big(f_{\theta_e}(\tilde x',\tilde m',a_{t-K+1:t}),(\bar a_h)_{h\in\mathcal H}\big)$
\State $\mathcal L_{\mathrm{schema}}$ by \eqref{eq:lsch}
\algcomment{one-sided: only the reduced view is pulled toward $\mathrm{sg}[\hat z_{t+h}]$}
\State $\mathcal L_{\mathrm{SSL}}$ by \eqref{eq:lssl}
\State update $(\theta_e,\phi,\omega)$ by AdamW on $\nabla\mathcal L_{\mathrm{SSL}}$, gradient norm clipped to $1$
\algcomment{the physical head $\psi$ is not used in this stage}
\State $\bar\theta_e \gets \tau\bar\theta_e+(1-\tau)\theta_e$
\algcomment{after the optimiser step}
\end{algorithmic}
\end{algobox}

The locked candidate drops each channel for the whole context window with probability $p_{\mathrm{drop}}=0.15$ and sets $\kappa=0.65$, $\tau=0.996$ and $(\lambda_{\mathrm{lat}},\lambda_{\mathrm{vic}},\lambda_{\mathrm{sch}},\lambda_{\mathrm{act}})=(1.0,\,0.05,\,0.10,\,0.05)$; the complete configuration is listed in \hyperref[app:config]{App.~G}. Checkpoints of this stage are selected on a deterministic held-out SSL validation pass that scores the latent and VICReg terms only, never on the training loss. The second stage attaches a fresh probabilistic head and fine-tunes the whole network jointly, adding the Gaussian negative log-likelihood of the physical output to the same self-supervised terms, with schema consistency moved to the physical output; checkpoints of this stage are selected on source-validation RMSE (Sec.~\ref{sec:head}).

\subsection{Physical forecasting head and adaptation}\label{sec:head}
A probabilistic head with parameters $\psi$ maps each predicted latent to a diagonal Gaussian over the channels,
\begin{equation}\label{eq:head}
p_\psi(x_{t+h}\mid\hat z_{t+h})=\mathcal N\big(\mu_\psi(\hat z_{t+h}),\,\mathrm{diag}\,\sigma^2_\psi(\hat z_{t+h})\big),
\end{equation}
for $h\in\mathcal H$, with the likelihood evaluated on present channels only. Composed with the encoder and the predictor, it gives a factorized model of the marginals of \eqref{eq:target} on the grid $\mathcal H$, with $\theta=(\theta_e,\phi,\psi)$ ($r_\omega$ is auxiliary and not part of $\theta$):
\begin{equation}\label{eq:chain}
\begin{split}
p_\theta(x_{t+\mathcal H}\mid\cdot)&=\textstyle\prod_{h\in\mathcal H}p_\psi(x_{t+h}\mid\hat z_{t+h}),\\
\hat z_{t+h}&=\big[g_\phi\big(z^c_t,(\bar a_{h'})_{h'\in\mathcal H}\big)\big]_h,
\end{split}
\end{equation}
with $z^c_t$ given by \eqref{eq:enc}. Two remarks qualify \eqref{eq:chain}. (i) Pretraining fits only the latent factor: the physical head has weight zero and receives no gradient, and \eqref{eq:target} is reached only in the second stage, when the Gaussian negative log-likelihood of \eqref{eq:chain} is added to the same self-supervised terms, with schema consistency moved to the physical output. $\mathcal L_{\mathrm{SSL}}$ is thus a surrogate for \eqref{eq:target}, not a bound on its likelihood, and whether it helps is an empirical question; Sec.~\ref{sec:pretrain-results} finds no clean-source gain. (ii) The product in \eqref{eq:chain} is a joint law that treats the horizons, and, with the diagonal head, the channels, as conditionally independent given the context and the commands; it models the per-horizon marginals of \eqref{eq:target} but not their correlations, and the future commands enter only through their means $\bar a_h$.

The primary fine-tuning protocol attaches a fresh head, because a head present during pure SSL has had no supervised physical target. Three matched controls are retained: full training from scratch, pretrained body with a fresh head, and complete checkpoint transfer.

Target adaptation updates a restricted subset of parameters using target support windows. Zero-shot transfer uses the source-trained model without target updates. Few-shot transfer takes the leading fraction of each target run as support, updates the predictor and the physical head for 50 steps at learning rate $10^{-4}$, and evaluates on the remaining, later windows of the same runs, so support and query windows are disjoint. Online adaptation is reserved for causal prequential evaluation and is not used in the results reported in this study.

\subsection{Instance-normalized variant}\label{sec:revin}
The locked model normalizes every channel with a fixed source-train z-score. The official forecasting baselines instead apply RevIN \citep{kim2022revin}, which centers and scales each window by its own statistics and inverts the transform on the output. To test whether the zero-shot gap of Sec.~\ref{sec:revin-results} is a normalization effect rather than a property of the latent objective, a variant of the model applies a presence-aware RevIN inside the network with the settings of the official baselines (no affine parameters, no last-value subtraction, $\epsilon=10^{-5}$), so that the two arms differ by this transform only and by no parameter. For each window and channel $k$, the center $c_k$ and scale $s_k$ are computed on the $K$ context rows only and on present entries only, $c_k=\sum_t m_{t,k}x_{t,k}/\sum_t m_{t,k}$ and $s_k^2=\sum_t m_{t,k}(x_{t,k}-c_k)^2/\sum_t m_{t,k}+\epsilon$, and are detached from the graph; masked and absent entries are stored as zeros and would bias a plain mean, a choice shared with the official SimMTM pretraining code. A channel never observed in a window keeps the identity transform, so the seven channels absent on the target machine behave exactly as in the locked model. The target rows $x_{t+\mathcal H}$ are normalized with the context statistics before entering $f_{\bar\theta_e}$, so no future statistic reaches the model and every latent loss is computed in instance space. The physical head predicts in instance space, and its mean and log-variance are mapped back to the global z-space, $\hat\mu_k\leftarrow s_k\hat\mu_k+c_k$ and $\log\hat\sigma_k^2\leftarrow\log\hat\sigma_k^2+2\log s_k$, before any loss or metric, so the evaluation space, the trivial-predictor anchor and the adaptation procedure are unchanged. With the transform disabled the code path is bitwise that of the locked model, which regression tests enforce.

\begin{figure*}[t]
  \centering
  \includegraphics[width=0.92\textwidth]{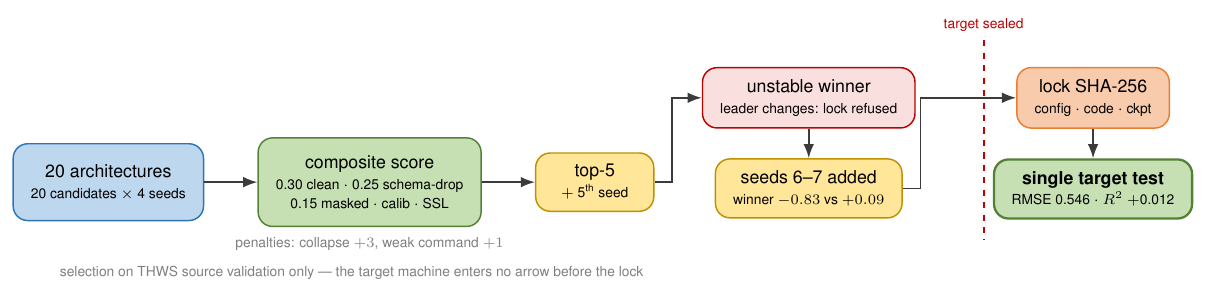}
  \caption{\textbf{Leakage-free model-selection pipeline.}
  Twenty pre-defined candidates are ranked on THWS source validation only, first with four seeds and then with a fifth seed for the top-5. When the provisional winner changes after confirmation, locking is refused and extra seeds are added to the leading candidates. In the present study, the per-horizon VICReg variant led the discovery stage, but the schema-consistency variant with command recovery remained superior after the full stability check and was frozen before a single confirmatory evaluation on the JOANNEUM target machine.}
  \label{fig:protocol}
\end{figure*}

\section{Experimental Protocol}\label{sec:protocol}
\subsection{Data and audits}
The source machine is the Spinner U5-620 five-axis machining center of the THWS production dataset with multiple changeovers \citep{martinez2025thws}; the target machine is provided by the FH JOANNEUM CNC machining data repository \citep{brillinger2025joanneum}. Both datasets are public under CC BY 4.0. The merged table (Table~\ref{tab:data}) contains 1,533,700 rows and 25 columns. THWS contributes 29,785 samples at 1 Hz, segmented into 62 NC-program sessions after excluding jog periods and splitting at temporal gaps longer than 5 s. JOANNEUM contains 1,503,915 raw controller-cycle records across 7 runs and 13 sessions. The source provides 17 canonical sensor channels, whereas 10 current/torque/power channels are shared with the target. Four mapped command variables represent spindle speed and the commanded X/Y/Z feeds.

Source sessions are divided into 42 training, 11 validation, and 9 test sessions with group-disjoint splitting. The resulting source window counts are 18,267 / 3,423 / 5,189, with zero window overlap across partitions. Windows are indexed inside each (machine, session, run) group and never cross a group boundary, so a group must contain at least $K+\max\mathcal{H}=48$ rows before it yields a single window, whose context is the $K=32$ rows ending at $t$ and whose targets are the five rows $t+h$, $h\in\mathcal{H}$. The same requirement explains why the 5\% support subset of Fig.~\ref{fig:fewshot} contains no complete window. All normalizers are fitted on source training data only. A unit audit detected three mismatches that were corrected deterministically before modeling: JOANNEUM power channels were stored in watts while THWS used kilowatts (factor $10^{-3}$); commanded axis feeds were recorded in mm/s on one machine and mm/min on the other (factor 60); and spindle speed was logged in degrees per second on one machine and rpm on the other (factor 6, verified from position derivatives). The last two matter because the command channels condition the predictor. Target statistics are not used to fit source normalizers. Validation masking is deterministic, DataLoader workers are seeded, checkpoint loading reports missing and unexpected keys, and regression tests cover split integrity, masking determinism, EMA and stop-gradient behavior, and metric aggregation. All experiments ran on one NVIDIA DGX Spark (one GB10 GPU), about six GPU-days in total.

\begin{table}[t]
\centering
\caption{Audited cross-machine data protocol.}
\label{tab:data}
\footnotesize
\setlength{\tabcolsep}{4pt}
\begin{tabular}{@{}lcc@{}}
\toprule
\rowcolor{cvprblue!10}
 & THWS source & JOANNEUM target \\
\midrule
Raw rows & 29,785 & 1,503,915 \\
Independent units & 62 sessions & 7 runs / 13 sessions \\
Working rate & 1 Hz & 1 Hz (from 2 ms cycles) \\
Canonical/shared sensors & 17 & 10 shared \\
Commands & \multicolumn{2}{c}{spindle + commanded X/Y/Z feeds} \\
Normalization & source-train only & source normalizer \\
\bottomrule
\end{tabular}
\end{table}

\subsection{Model selection and baselines}\label{sec:selection}
The source-only architecture search defines 20 candidates prior to execution. The search spans masking strategies, command-injection mechanisms, VICReg placement, latent-loss weight, horizon aggregation, schema consistency, command recovery, EMA dynamics, model capacity, normalization, and head initialization. Each candidate is first run with four seeds. Ranking uses a composite source-validation score: each component is z-scored within the candidate pool and combined as $0.30$ clean RMSE $+\,0.25$ schema-drop RMSE $+\,0.15$ masked RMSE $+\,0.10$ long-horizon RMSE $+\,0.10$ calibration penalty $+\,0.10$ SSL validation loss, plus penalties of $3.0$ per collapsed seed and $1.0$ for weak command usage (lower is better). The top-5 candidates receive a fifth seed. If the provisional winner changes under confirmation, model locking is refused and additional seeds are added to the tied leaders. This occurred here: the per-horizon VICReg variant led the discovery stage, but the schema-consistency variant with command recovery overtook it after confirmation and remained superior after seven valid seeds, after which the configuration, code, normalizer, and checkpoint were frozen before a single confirmatory pass on the JOANNEUM target machine. A pre-lock model was also evaluated on the target (Sec.~\ref{sec:fewshot}); those reads did not enter the ranking or the lock, which use source validation only.

The six components are measured on source validation after fine-tuning: clean RMSE on the unmasked source-validation windows; schema-drop RMSE, obtained with the schema restricted to the 10 channels shared with the target; masked RMSE under deterministic corruption masks; long-horizon RMSE, taken as the worst of the five per-horizon RMSE values; a calibration penalty, the absolute difference between the empirical coverage at the nominal 90\% level and 0.90; and the best held-out SSL validation loss reached during pretraining. Each component is z-scored within the pool of usable runs, a component with zero variance across that pool contributing 0, and the resulting score is averaged over the seeds of a candidate. Algorithm~\ref{alg:lock} states the procedure; the locked values are listed in \hyperref[app:config]{App.~G}.

\refstepcounter{algorithm}\label{alg:lock}
\begin{algobox}{Algorithm \thealgorithm: Source-only model selection and lock}
\small
\begin{algorithmic}[1]
\Require candidate set $\mathcal C$, $|\mathcal C|=20$, written down before any run; source train/validation splits; component weights $w$; penalties $\pi_{\mathrm{col}}=3$, $\pi_{\mathrm{act}}=1$; the target machine stays sealed
\For{each $c\in\mathcal C$ and each of four seeds}
  \State train $c$ (Algorithm~\ref{alg:train}, then head fine-tuning); measure the six components on source validation
  \State drop the run from the pool if the pretraining latents collapse (latent std $<0.05$ or effective rank $<10\%$ of $d$); flag post-fine-tuning collapse and weak command use ($\mathrm{RMSE}(\text{shuffled }\bar a)/\mathrm{RMSE}(\text{true }\bar a)<1.02$)
\EndFor
\State z-score each component within the pool of valid runs; per seed $S\gets \sum_k w_k z_k + \pi_{\mathrm{col}}\,\mathbf 1[\text{collapsed}] + \pi_{\mathrm{act}}\,\mathbf 1[\text{weak}]$; $S(c)\gets$ mean over the valid seeds of $c$
\State $\mathcal T\gets$ the five lowest $S$; add a fifth seed to each $c\in\mathcal T$; recompute $S$ within $\mathcal T$
\While{the leader of $\mathcal T$ changed since the previous seed count}
  \State refuse to lock; add seeds to the leading candidates; recompute $S$
  \algcomment{seven seeds settled it here}
\EndWhile
\State $c^\star\gets\arg\min_{c\in\mathcal T} S(c)$; lock only if $c^\star$ has no collapse flag, no weak-command flag, at least five valid seeds, a stable rank-1 across the confirmation stage, and the confirmation stage was run; freeze configuration, code, normalizer and the best-composite checkpoint with SHA-256 hashes
\State verify the hashes and evaluate $c^\star$ \textbf{once} on the sealed target windows
\algcomment{the only pass of the locked model on the target; the post-lock ablation below is declared separately}
\end{algorithmic}
\end{algobox}

\paragraph{Post-lock declared ablation.} The lock consumed the single declared target pass. One paired ablation was run afterwards, in response to a reviewer remark that the model lacked the instance normalization used by the target baselines. Its protocol was fixed before any run: the control arm re-uses the locked configuration (\hyperref[app:config]{App.~G}) verbatim, and its three re-runs reproduce the source-validation metrics of the corresponding search seeds exactly; the variant arm differs only by the RevIN flag (Sec.~\ref{sec:revin}), both arms are trained with the same three seeds through the same two-stage pipeline, and a second, declared opening of the target windows was conditioned on the variant winning all four RMSE components of the composite score on source validation. That verdict was recorded before the target stage ran, and the target stage then evaluated all six training runs (3 seeds $\times$ 2 arms) with no selection on the target. Besides the diagnostic reads by the pre-lock model (Sec.~\ref{sec:fewshot}), the target windows have therefore been read twice: once by the locked model, which remains the only confirmatory result, and once by this ablation, whose numbers are reported as diagnostic evidence with their seed count. The lock itself was not revisited.

Baselines include supervised MLP, GRU, LSTM, TCN, DLinear, Transformer, TSMixer, and RSSM models, together with trivial persistence and linear-drift references. For the target comparison, the official PatchTST and iTransformer repositories are trained once on the same source-train windows (single seed, repository defaults, RevIN enabled) and evaluated on the same sealed target windows; they receive the 17 sensors plus the 4 command channels as input, and channels absent on the target are zero-imputed in normalized space (Appendix~E). Six in-house reconstruction-pretraining baselines (MAE-, SimMTM-, PatchFormer-, EMIT-, MMR- and LoMaR-style) were implemented but are not reported here; the official SimMTM repository did not converge on this data regime.

\subsection{Metrics}
Forecasting is evaluated using RMSE, MAE, $R^2$, and per-horizon/per-channel errors. The probabilistic head is evaluated with Gaussian NLL and empirical predictive coverage. Where target inferential statistics are reported, the 7 JOANNEUM runs are the independent units; windows are not used as independent replicates. $R^2$ is computed against the pooled mean of the evaluated channels and horizons.

\section{Results}
\subsection{Corrected JEPA pretraining is stable but does not improve clean-source RMSE}\label{sec:pretrain-results}
A protocol audit identified three issues in the initial pretraining implementation: checkpoint selection based on training loss, latent schema consistency applied to an untrained physical head, and representational collapse in the EMA target encoder. The corrected pipeline selects checkpoints on deterministic held-out SSL validation, moves schema consistency to latent space, audits checkpoint loading, and applies VICReg to the pooled context representation (the target half of \eqref{eq:lvic} is monitored only; the target encoder inherits the effect through the EMA). Target-encoder effective rank increased from approximately 5\% to 58\% of the latent dimensionality after correction. Effective rank is defined here as the exponential of the entropy of the singular values of the centred latent matrix after normalizing them to sum to one, and is reported as a fraction of the latent dimensionality $d$; a run counts as collapsed when the smallest per-dimension standard deviation falls below 0.05 or the effective rank falls below 10\% of $d$.

\begin{table*}[t]
\centering
\caption{\textbf{Source-validation results.} (a) Matched five-seed comparison of initializations (mean $\pm$ SD over seeds; lower is better; trivial baselines here and in Fig.~\ref{fig:baselines} come from two separate evaluation scripts and differ slightly). (b) Selected ablations, single seed each on the pre-search base configuration; best tested setting per factor.}
\label{tab:p3ablation}
\small
\begin{subtable}[t]{0.46\linewidth}
\centering
\caption{Initialization (five seeds).}
\label{tab:p3}
\begin{tabular}{lc}
\toprule
\rowcolor{cvprblue!10}
Model initialization & RMSE \\
\midrule
Scratch & $0.811\pm0.022$ \\
Pretrained body + fresh head & $0.813\pm0.022$ \\
Complete pretrained checkpoint & $0.812\pm0.012$ \\
\midrule
Linear drift & 0.928 \\
Persistence & 1.135 \\
\bottomrule
\end{tabular}
\end{subtable}\hfill
\begin{subtable}[t]{0.50\linewidth}
\centering
\caption{Ablations (single seed).}
\label{tab:ablation}
\begin{tabular}{lll}
\toprule
\rowcolor{cvprblue!10}
Factor & Best tested setting & RMSE \\
\midrule
Command injection & token & 0.821 \\
Mask ratio & 0.03 & 0.804 \\
Latent weight & $\lambda_{\mathrm{lat}}=0.25$ & 0.774 \\
VICReg & per-horizon, 0.1 & 0.798 \\
Transformer norm & pre-norm & 0.829 \\
Horizon aggregation & pooled & 0.787 \\
EMA & 0.9995 & 0.799 \\
\bottomrule
\end{tabular}
\end{subtable}
\end{table*}

A matched five-seed comparison isolates the effect of pretraining (Table~\ref{tab:p3}). Training from scratch yields $0.811\pm0.022$ RMSE, pretrained body with a fresh head yields $0.813\pm0.022$, and complete checkpoint transfer yields $0.812\pm0.012$. The scratch-versus-pretraining difference is not significant ($p=0.76$). The scratch arm is trained with the same self-supervised terms in stage 2; it removes the pretraining stage, not the latent objective. Persistence and linear drift remain substantially weaker. Thus, no claim of improved clean-source forecasting from JEPA pretraining is supported by the current evidence.

\subsection{Representation hyperparameters materially affect stability and forecasting}\label{sec:hparams}
The latent-loss weight has a pronounced effect on downstream forecasting (Table~\ref{tab:ablation}). The best tested value is $\lambda_{\mathrm{lat}}=0.25$ with RMSE 0.774; larger values degrade performance to 0.821 at $\lambda_{\mathrm{lat}}=1$ and 0.845 at $\lambda_{\mathrm{lat}}=4$. VICReg improves clean-source RMSE from 0.815 without regularization to 0.798 with per-horizon weighting and is additionally required to prevent collapse. Pooled horizon aggregation reaches 0.787 compared with 0.829 for per-horizon aggregation. Slow EMA targets are also preferable, with 0.9995 reaching 0.799 versus 0.808 at 0.999 and approximately 0.83 for faster updates. Two further observations temper the picture: removing the latent loss entirely ($\lambda_{\mathrm{lat}}=0$) yields 0.787, so the latent prediction objective at its default weight costs clean-source accuracy, and the best clean RMSE among mask ratios (0.804) is obtained with the lightest ratio ($\rho=0.03$). In the mixed mode of these ablations $\rho$ scales the point, block and event components only; whole-channel dropout at probability 0.15 is applied at every ratio, so even the lightest setting removes whole sensors. All ablations are single-seed runs on the pre-search base configuration. The locked candidate nevertheless keeps $\lambda_{\mathrm{lat}}=1$, per-horizon aggregation and EMA 0.996, and masks by whole-channel dropout alone ($p_{\mathrm{drop}}=0.15$; $\rho$ plays no role in this mode), because the composite selection score (Sec.~\ref{sec:selection}) weights schema-drop and masked robustness, calibration and command usage in addition to clean RMSE, and the discovery grid did not combine the ablation optima with schema consistency and command recovery. These ablations indicate that JEPA stability on CNC time series depends strongly on representation-objective balance rather than on a single default configuration; LeWorldModel \citep{maes2026lewm} removes most of these weights by construction, which is a direct route to reducing this sensitivity.

\begin{figure*}[t]
  \centering
  \includegraphics[width=0.9\textwidth]{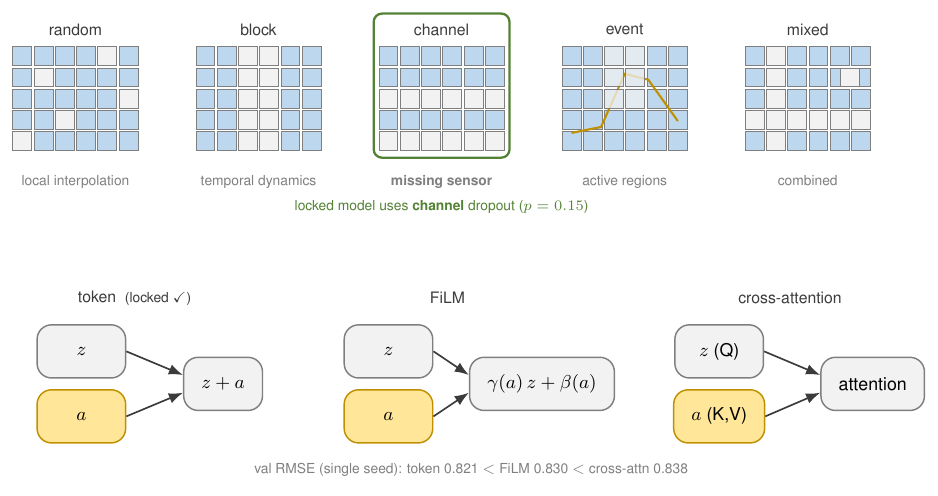}
  \caption{\textbf{Masking regimes and command-injection variants spanned by the search.} Top: the five training-time corruption modes (random points, temporal blocks, whole channels, high-activity events, and their union); the locked candidate uses channel masking, which removes entire sensors from the context and mimics a channel absent on the target machine. Bottom: the three command-injection mechanisms of the predictor. Single-seed runs on the pre-search base configuration; the token entry is the one reported in Table~\ref{tab:ablation}: token 0.821, FiLM 0.830, cross-attention 0.838.}
  \label{fig:masking}
\end{figure*}

The search also spans the corruption applied to the context window and the mechanism by which the command enters the predictor (Figure~\ref{fig:masking}). Five masking modes are available: random masking removes individual (time, channel) entries; block masking removes contiguous runs of 2 to 12 steps inside a channel; channel masking removes whole sensors from the window, each with a fixed probability $p_{\mathrm{drop}}$ that does not depend on the ratio; event masking removes a random subset, sized by the mask ratio, of the samples whose step-to-step change exceeds a per-channel quantile of the window; and the mixed mode takes the union of the four. The locked candidate uses channel masking, which removes entire sensors from the context and therefore mimics a channel that is absent on the target machine. Three command-injection mechanisms are compared: token injection adds a projection of the mean future command to each horizon slot, FiLM applies a feature-wise affine modulation of the context representation derived from that command, and cross-attention conditions the horizon slots on the command; these are single-seed runs on the pre-search base configuration, and the token entry is the one reported in Table~\ref{tab:ablation}. Source-validation RMSE is 0.821 for token, 0.830 for FiLM and 0.838 for cross-attention, and the locked candidate uses token injection.

\subsection{Source-only model selection identifies a schema-consistent command-recovery variant as the locked configuration}
The architecture search separates useful design choices from failure modes. The plain JEPA control, which removes anti-collapse regularization, collapses on 4/4 seeds and is excluded from ranking. Large-model scaling is also unstable in this data regime: the largest candidate ($d{=}512$, eight layers) collapses on 3/4 seeds. In contrast, the selected variant combines latent schema consistency and command recovery while preserving clean-source performance. Its source-validation score is $0.8216\pm0.0089$ RMSE with a schema-drop score among the strongest of the valid candidates (0.918) and clear command sensitivity (1.058 under command shuffling); its selection reflects the composite score rather than any single metric.

The stability rule is consequential. The per-horizon VICReg variant ranks first at the four-seed discovery stage, but the schema-consistency variant with command recovery overtakes it after five and then seven seeds. Locking on the initial discovery result would therefore have selected the wrong configuration. Figure~\ref{fig:protocol} summarizes the sealed selection pipeline, which keeps the JOANNEUM target data outside the ranking: no target metric, including those of the pre-lock model (Sec.~\ref{sec:fewshot}), enters the selection score.

\begin{figure*}[t]
  \centering
  \begin{subfigure}[t]{0.5\linewidth}
    \centering
    \includegraphics[width=\linewidth]{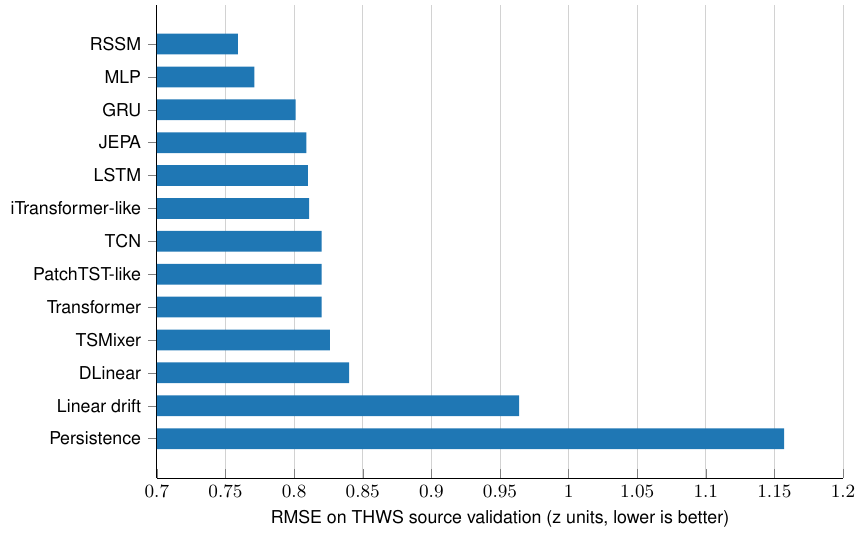}
    \caption{Clean-source comparison on THWS validation. ``-like'' bars are in-house adaptations; the JEPA bar is the pre-search reference model on the source test split, all other bars are source-validation values.}
    \label{fig:baselines}
  \end{subfigure}\hfill
  \begin{subfigure}[t]{0.46\linewidth}
    \centering
    \includegraphics[width=\linewidth]{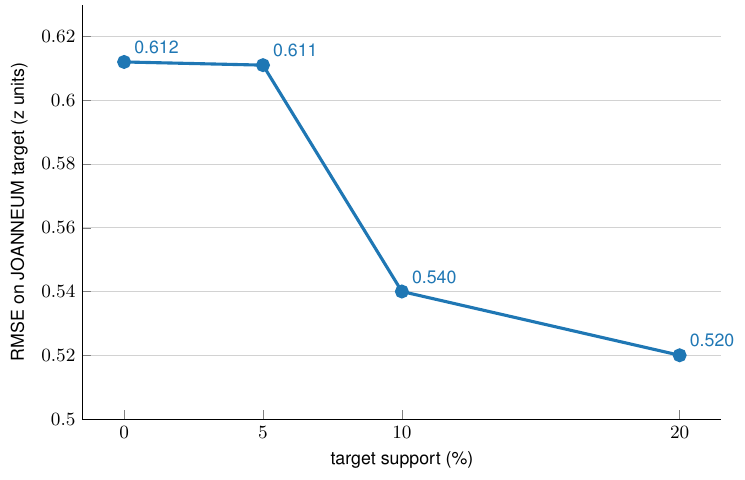}
    \caption{THWS$\rightarrow$JOANNEUM few-shot adaptation of the pre-lock model. The 5\% point is effectively zero-shot because the support subset holds no complete 48-step window.}
    \label{fig:fewshot}
  \end{subfigure}
  \caption{\textbf{Source forecasting benchmark (a) and pre-lock few-shot transfer (b).}}
  \label{fig:fig45}
\end{figure*}

\subsection{Strong forecasting baselines remain competitive on the source machine}
The clean-source forecasting benchmark is summarized in Fig.~\ref{fig:baselines}. RSSM obtains RMSE 0.759 and MLP 0.771, outperforming the pre-search JEPA reference model at 0.809 (source test split) and the locked candidate at 0.822 (source validation). GRU, LSTM, and iTransformer-like models are also competitive. All learned models outperform linear drift and persistence. The result rules out a generic source-forecasting superiority claim and motivates evaluation on cross-machine adaptation and partial sensor overlap instead.

\subsection{The locked model transfers zero-shot, but official zero-shot forecasters remain stronger in RMSE}
The confirmatory JOANNEUM evaluation is executed once, after locking the schema + command-recovery configuration on source data only. The locked model reaches zero-shot RMSE 0.546, $R^2=0.012$, and NLL 0.52 on the target machine, improving over persistence at 0.654 (Table~\ref{tab:target}). This establishes that the transferred model beats a trivial baseline on an unseen machine despite the reduction from 17 to 10 available channels. The evaluated checkpoint is the single best-composite seed among the seven source-validation seeds, so the target number is a single-seed result. Per horizon, target $R^2$ is 0.035, 0.055, 0.052, $-0.023$ and $-0.059$ at $h=1,2,4,8,16$: the transferred model explains a small share of variance at short horizons and falls below the pooled target-mean predictor at 8 and 16\,s.

However, official PatchTST and iTransformer baselines with RevIN \citep{kim2022revin} achieve lower zero-shot RMSE, 0.503 and 0.498, respectively. The claim must therefore remain specific: what the world model adds is a probabilistic output and a structure for command conditioning and adaptation, not state-of-the-art zero-shot RMSE; its command use and adaptation on the target remain to be verified on the locked model (Sec.~\ref{sec:limitations}). Section~\ref{sec:revin-results} shows that this margin disappears once the same instance normalization is applied inside the model, at a calibration cost.

\begin{table*}[t]
\centering
\caption{Confirmatory JOANNEUM zero-shot transfer after source-only model locking, and the post-lock RevIN ablation. Source column: official baselines are scored on the source test split (5,189 windows, 17 sensors); the world-model rows are the seven-seed (locked) and three-seed (RevIN) mean $\pm$ SD on source validation. Target columns: 10 shared channels, 2,457 windows. The locked row is the single declared target pass; the RevIN row is a second, declared read of all three seeds with no selection on the target (Sec.~\ref{sec:selection}). Lowest target RMSE in bold.}
\label{tab:target}
\footnotesize
\setlength{\tabcolsep}{4pt}
\begin{tabular}{lcccl}
\toprule
\rowcolor{cvprblue!10}
Method & Source RMSE & Target zero-shot RMSE & Target NLL & Notes \\
\midrule
Persistence & --- & 0.654 & --- & trivial floor \\
PatchTST (official, RevIN) & 0.804 & 0.503 & --- & deterministic, single seed \\
iTransformer (official, RevIN) & 0.822 & 0.498 & --- & deterministic, single seed \\
World model (locked) & $0.822\pm0.009$ & 0.546 & 0.52 & probabilistic, command-cond.; single declared pass \\
World model + RevIN (post-lock) & $0.766\pm0.001$ & $\mathbf{0.495\pm0.004}$ & 20.6 & probabilistic, command-cond.; 3 seeds, diagnostic \\
\bottomrule
\end{tabular}
\end{table*}

Only the target columns of Table~\ref{tab:target} are a like-for-like comparison: the same 2,457 target windows, the same 10 shared channels, and no target-side selection for any method; the source column mixes the official baselines' source test split with the world model's source validation, as the caption states, and the RevIN row is diagnostic rather than confirmatory.

\subsection{Instance normalization closes the zero-shot RMSE gap but breaks target calibration}\label{sec:revin-results}
The post-lock paired ablation of Sec.~\ref{sec:selection} tests whether the gap to the official baselines is a normalization effect. On source validation the RevIN variant wins on all three seeds for every RMSE component: clean RMSE $0.819\pm0.010\to0.766\pm0.001$, schema-drop $0.918\pm0.005\to0.887\pm0.003$, masked $0.827\pm0.010\to0.786\pm0.002$ and long-horizon $0.868\pm0.007\to0.849\pm0.004$, with bootstrap 95\% intervals of the paired difference excluding zero (\hyperref[app:revin]{App.~H}). The calibration penalty falls from 0.124 to 0.035, command sensitivity rises from 1.06 to 1.18, and the seed-to-seed spread shrinks by a factor of five to ten. With three paired seeds the exact sign-flip test cannot go below $p=0.125$, so these are read as consistent paired wins, not as significance claims.

On the target machine the three control seeds give RMSE $0.555\pm0.015$ (the locked seed is 0.546) and the three RevIN seeds $0.495\pm0.004$; the paired differences are $-0.046$, $-0.079$ and $-0.054$, and $R^2$ moves from $-0.02$ to $+0.19$ (Table~\ref{tab:target}). The variant is therefore level with official PatchTST (0.503) and iTransformer (0.498), which were run once each without hyperparameter search, so no state-of-the-art claim follows. The gain concentrates on the channels whose level shifts most across machines (Z-axis current RMSE 0.84 to 0.20), while spindle current and torque degrade slightly (0.37 to 0.43 and 0.60 to 0.67).

The price is calibration. Target NLL rises from 0.89 to 20.6 and empirical coverage at the nominal 90\% level falls from 0.953 to 0.874. On seed 0, 11.5\% of the RevIN predictions sit at the log-variance floor of the head ($-8$, against 0\% for the control), and the per-channel NLL is dominated by spindle torque (89 nats), Z power (51), spindle current (40) and spindle power (28). The mechanism is the degenerate case of instance normalization on stationary windows: when the spindle is stopped or a power channel is constant over the 32-step context, the window scale is $s_k=\sqrt\epsilon\approx3\times10^{-3}$, the term $2\log s_k$ then pushes the de-normalized log-variance far below any physical value, and a start-up inside the 16-step horizon costs hundreds of nats. RMSE is protected because the mean is de-normalized with the same scale. A latent-health gate also flags the three RevIN runs on its raw-scale criterion (minimum predicted-latent standard deviation 0.040--0.041 against a threshold of 0.05; control 0.059--0.061), whereas the scale-free criterion, effective rank, is twice as high with RevIN (0.45--0.49 against 0.22--0.23) and latent health after fine-tuning is identical; the gate was left unchanged and must become scale-free before RevIN enters any future selection.

The reading for world models is direct. Input normalization is sufficient to explain the zero-shot RMSE advantage of the official forecasters, so that advantage is not by itself evidence against the command-conditioned latent objective: the same objective with the same normalization matches them while keeping the command input and the probabilistic head, though not its calibration. Whether the objective helps on the target is a separate question, which no arm here answers (Sec.~\ref{sec:limitations}). But pure RevIN, as the baselines use it, is incompatible with a Gaussian head under stationary regimes, which is precisely where an industrial world model must remain uncertainty-aware. The locked model is therefore kept as the reference result and the variant is reported with its cost.

\subsection{Few-shot adaptation remains promising but is still diagnostic}\label{sec:fewshot}
A pre-lock adaptation sweep suggests that limited target support can further improve transfer (Figure~\ref{fig:fewshot}). On the evaluated pre-lock model, zero-shot transfer yields RMSE 0.612, MAE 0.401, and $R^2=-0.244$. With 10\% target support, RMSE decreases to 0.540 and $R^2$ becomes positive (0.046). With 20\% support, RMSE reaches 0.520, MAE 0.299, and $R^2=0.167$. The 5\% support setting does not contain a valid 48-step support window and therefore reduces to the zero-shot model rather than constituting an adaptation failure.

\section{Discussion}
The evidence supports a narrow interpretation. First, JEPA pretraining does not improve clean-source forecasting over matched scratch training. Second, anti-collapse regularization and source-only selection stability are not minor implementation details: the plain JEPA control collapses completely, large scaling is unstable, and the provisional winner changes during confirmation. Third, cross-machine evaluation provides a distinct axis from in-domain forecasting. The locked schema + command-recovery model transfers zero-shot and outperforms persistence on an unseen machine, even though stronger zero-shot RMSE is currently achieved by official RevIN-equipped forecasting baselines. Fourth, a three-seed diagnostic ablation attributes that RMSE margin to input normalization: the same architecture with RevIN matches the baselines but loses its calibration on stationary windows, so normalization and uncertainty must be designed together rather than borrowed from point forecasters.

Industrial predictive representations should therefore be judged on whether a learned state responds to the commands, uncertainty-aware, robust to schema shift, and adaptable under limited target data, not on source interpolation accuracy alone.

\paragraph{Future work.} A second version of the study (V2, \hyperref[app:concept]{App.~F}) should (i) train on more than one source machine so that invariance to machine-specific dynamics is measured rather than assumed; (ii) make instance normalization compatible with the probabilistic head, by flooring the window scale in z-units, blending to the identity below a context-variance threshold, or keeping pure RevIN for the mean while predicting the variance in global space, each of which departs from the baselines' pure RevIN and must be reported as such, and make the latent-health gate scale-free before the twenty-candidate search is repeated with normalization enabled; (iii) connect the latent state to planning or model-based RL so it is judged by decision quality; and (iv) evaluate policies off-line before any real-machine test. Causal representation learning stays outside the claim set until the data provide enough intervention diversity (Fig.~\ref{fig:v2}, \hyperref[app:concept]{App.~F}).

\section{Limitations}\label{sec:limitations}
One source and one target machine limit claims of general cross-machine generalization, and the 7 independent JOANNEUM runs restrict statistical power. The target stream is interpreted under a 2 ms controller-cycle assumption, and sensitivity to resampling remains a separate robustness analysis; seven source-only channels have no target counterpart. The few-shot curve was obtained before the model lock and must be re-measured on the locked configuration before being treated as confirmatory. The confirmatory target result is a single checkpoint evaluated once; the target windows were read by the pre-lock model, once by the locked model, and once by the declared RevIN ablation, so no further target evaluation of this pipeline can be called sealed; that ablation has three paired seeds, its runs carry a raw-scale latent-health flag, and its calibration collapse is diagnosed but not fixed. Run-level confidence intervals and command-sensitivity tests on the target machine have not yet been computed for the locked model. On the pre-lock model, zeroing or shuffling future commands changed target RMSE by less than 0.005, so command use under transfer is unverified and is the first item to test; that model was also under-calibrated (67\% empirical coverage at nominal 90\%). Every evaluated arm, including scratch training, is trained with the latent objective, and no target arm without the latent objective has been evaluated, so the transfer results do not isolate its contribution. The intensity of whole-channel masking was not varied either: in channel mode the code ignores the ratio, so the ``heavy channel masking'' candidate of the search ran with the same drop probability (0.15) as the locked model, and its near-identical scores (App.~\ref{app:search}) reflect that. The ablations in Table~\ref{tab:ablation} are single-seed, and the official baselines are single-seed runs without hyperparameter search. Finally, the present paper does not yet evaluate downstream policy learning or off-policy control, so the world-model interpretation remains predictive rather than decision-theoretic.

\section{Conclusion}
This study presents a schema-adaptive, command-conditioned JEPA with an audited cross-machine CNC protocol under partial sensor overlap. JEPA pretraining does not improve clean-source RMSE over matched scratch training; source-only search still identifies a stable schema-consistent, command-recovery configuration that transfers zero-shot to an unseen machine above persistence, while official RevIN-based forecasters remain stronger in raw zero-shot RMSE; a post-lock paired ablation shows that instance normalization is sufficient to close that margin, at the cost of target calibration. The contribution is a controlled benchmark and methodology for industrial predictive representations under joint machine shift, schema shift, and command conditioning, not a state-of-the-art claim.

\section*{Code and data availability}
Code, configuration files, the twenty candidate specifications, and the audit scripts are available at \url{https://github.com/ostertagmatthieu-dev/saac-jepa}; animated versions of the schematics are on the project page, \url{https://ostertagmatthieu-dev.github.io/saac-jepa/}. The source and target datasets are public under CC BY 4.0: THWS five-axis CNC milling (\url{https://doi.org/10.5281/zenodo.14094887}) and the FH JOANNEUM CNC machining repository (\url{https://doi.org/10.17632/gtvvwmz7r7.2}).

\clearpage
\maketitlesupplementary
\appendix
\section{Architecture Search: Twenty Candidates}\label{app:search}
\small
The architecture search contains 20 pre-defined variants of the model. Candidate ranking used \textbf{THWS source validation only}; the JOANNEUM target data were excluded from candidate ranking. The cards below show the common computational skeleton and highlight the mechanism changed by each candidate. Metrics are means over valid seeds from the architecture-search log. The final stability check reported in the closure log selected \textbf{the schema-consistency variant with command recovery} after seven valid seeds for the two leading candidates, after which code, configuration, normalizers, and checkpoint were frozen before the confirmatory target pass.

\textbf{Reading the cards.} Each diagram follows \emph{masked sensor context $\rightarrow$ context encoder $\rightarrow$ command-conditioned predictor $\rightarrow$ future latent}, with an EMA target branch used for latent self-supervision. Tags report the masking regime, command-injection mechanism, latent objective, schema/command auxiliary losses, EMA schedule, capacity, normalization, and fine-tuning head initialization. Command sensitivity is $\mathrm{RMSE}(\text{shuffled commands})/\mathrm{RMSE}(\text{true commands})$; values above one indicate measurable command usage.

\vspace{2mm}
\noindent\begin{minipage}[t]{\linewidth}\centering
\begin{tikzpicture}\node[draw=mredL,rounded corners=6pt,line width=.8pt,inner sep=6pt,text width=.94\linewidth] {%
\begin{minipage}{\linewidth}\raggedright
\textbf{\large Plain JEPA control}\hspace{0.6em plus 1fill}\pill{mredL}{COLLAPSE 4/4}\\[-1mm]
{\fontsize{7.2}{8.0}\selectfont\color{mgrayL}INVALID: SSL LATENT COLLAPSE (4/4)}
\vspace{1.5mm}
\centering\resizebox{\linewidth}{!}{\begin{tikzpicture}[node distance=2mm]
\node[lat,minimum width=14mm] (x) {Sensors\\\textbf{Mixed 35\%}};
\node[enc,minimum width=16mm,right=2mm of x] (e) {Encoder\\$d=256$, $L=6$};
\node[prd,minimum width=16mm,right=2mm of e] (p) {Predictor\\\textbf{Token}};
\node[lat,minimum width=9mm,right=2mm of p] (zh) {$\hat z^{+}$};
\node[act,minimum width=14mm,below=3mm of p] (ac) {future commands};
\node[tgt,minimum width=16mm,below=3mm of e] (te) {EMA target\\0.996 fixed};
\draw[ar] (x)--(e); \draw[ar] (e)--(p); \draw[ar] (p)--(zh); \draw[ar] (ac)--(p); \draw[ema] (e)--(te);
\end{tikzpicture}}
\raggedright\vspace{1mm}
\pill{mblueL}{mask: Mixed 0.35} \pill{myellL}{command: Token} \pill{mgreenL}{VICReg: None} \pill{mgrayL}{place: -}\\[1mm]
\pill{mgrayL}{$\lambda_{lat}$=1.0} \pill{mgreenL}{$\lambda_{sch}$=0.0} \pill{myellL}{$\lambda_{act}$=0.0} \pill{mblueL}{target: Per-horizon} \pill{mgrayL}{latent LN: off}\\[1mm]
\pill{mpinkL}{EMA: 0.996 fixed} \pill{mblueL}{norm: z-score} \pill{mgrayL}{head: Fresh}
\vspace{1.2mm}
{\fontsize{6.9}{7.6}\selectfont \textbf{Source validation:} No valid source metrics; candidate excluded before ranking.\\\textbf{Selection:} Discovery exclu; confirmation exclu; invalid: ssl\_latent\_collapse.}
\end{minipage}};\end{tikzpicture}\end{minipage}\par\vspace{2.4mm}\noindent\begin{minipage}[t]{\linewidth}\centering
\begin{tikzpicture}\node[draw=mblueL,rounded corners=6pt,line width=.8pt,inner sep=6pt,text width=.94\linewidth] {%
\begin{minipage}{\linewidth}\raggedright
\textbf{\large JEPA + per-horizon VICReg}\hspace{0.6em plus 1fill}\pill{mblueL}{7/7 seeds}\\[-1mm]
{\fontsize{7.2}{8.0}\selectfont\color{mgrayL}VALID}
\vspace{1.5mm}
\centering\resizebox{\linewidth}{!}{\begin{tikzpicture}[node distance=2mm]
\node[lat,minimum width=14mm] (x) {Sensors\\\textbf{Mixed 35\%}};
\node[enc,minimum width=16mm,right=2mm of x] (e) {Encoder\\$d=256$, $L=6$};
\node[prd,minimum width=16mm,right=2mm of e] (p) {Predictor\\\textbf{Token}};
\node[lat,minimum width=9mm,right=2mm of p] (zh) {$\hat z^{+}$};
\node[act,minimum width=14mm,below=3mm of p] (ac) {future commands};
\node[tgt,minimum width=16mm,below=3mm of e] (te) {EMA target\\0.996 fixed};
\draw[ar] (x)--(e); \draw[ar] (e)--(p); \draw[ar] (p)--(zh); \draw[ar] (ac)--(p); \draw[ema] (e)--(te);
\end{tikzpicture}}
\raggedright\vspace{1mm}
\pill{mblueL}{mask: Mixed 0.35} \pill{myellL}{command: Token} \pill{mgreenL}{VICReg: Per-horizon} \pill{mgrayL}{place: ctx+tgt}\\[1mm]
\pill{mgrayL}{$\lambda_{lat}$=1.0} \pill{mgreenL}{$\lambda_{sch}$=0.0} \pill{myellL}{$\lambda_{act}$=0.0} \pill{mblueL}{target: Per-horizon} \pill{mgrayL}{latent LN: off}\\[1mm]
\pill{mpinkL}{EMA: 0.996 fixed} \pill{mblueL}{norm: z-score} \pill{mgrayL}{head: Fresh}
\vspace{1.2mm}
{\fontsize{6.9}{7.6}\selectfont \textbf{Source validation:} Clean 0.8336 $\pm$ 0.0033 | schema 0.9494 | cmd 1.0350 | eff.rank 0.3189\\\textbf{Selection:} Discovery \#1 (-0.346); confirmation \#2 (0.093).}
\end{minipage}};\end{tikzpicture}\end{minipage}\par\vspace{2.4mm}\noindent\begin{minipage}[t]{\linewidth}\centering
\begin{tikzpicture}\node[draw=mblueL,rounded corners=6pt,line width=.8pt,inner sep=6pt,text width=.94\linewidth] {%
\begin{minipage}{\linewidth}\raggedright
\textbf{\large Schema-adaptive latent consistency}\hspace{0.6em plus 1fill}\pill{mblueL}{4/4 seeds}\\[-1mm]
{\fontsize{7.2}{8.0}\selectfont\color{mgrayL}VALID}
\vspace{1.5mm}
\centering\resizebox{\linewidth}{!}{\begin{tikzpicture}[node distance=2mm]
\node[lat,minimum width=14mm] (x) {Sensors\\\textbf{Channel $p$=0.15}};
\node[enc,minimum width=16mm,right=2mm of x] (e) {Encoder\\$d=256$, $L=6$};
\node[prd,minimum width=16mm,right=2mm of e] (p) {Predictor\\\textbf{Token}};
\node[lat,minimum width=9mm,right=2mm of p] (zh) {$\hat z^{+}$};
\node[act,minimum width=14mm,below=3mm of p] (ac) {future commands};
\node[tgt,minimum width=16mm,below=3mm of e] (te) {EMA target\\0.996 fixed};
\draw[ar] (x)--(e); \draw[ar] (e)--(p); \draw[ar] (p)--(zh); \draw[ar] (ac)--(p); \draw[ema] (e)--(te);
\end{tikzpicture}}
\raggedright\vspace{1mm}
\pill{mblueL}{mask: Channel $p$=0.15} \pill{myellL}{command: Token} \pill{mgreenL}{VICReg: Per-horizon} \pill{mgrayL}{place: ctx+tgt}\\[1mm]
\pill{mgrayL}{$\lambda_{lat}$=1.0} \pill{mgreenL}{$\lambda_{sch}$=0.1} \pill{myellL}{$\lambda_{act}$=0.0} \pill{mblueL}{target: Per-horizon} \pill{mgrayL}{latent LN: off}\\[1mm]
\pill{mpinkL}{EMA: 0.996 fixed} \pill{mblueL}{norm: z-score} \pill{mgrayL}{head: Fresh}
\vspace{1.2mm}
{\fontsize{6.9}{7.6}\selectfont \textbf{Source validation:} Clean 0.827 $\pm$ 0.0058 | schema 0.9252 | cmd 1.0437 | eff.rank 0.3792\\\textbf{Selection:} Discovery \#11 (-0.272).}
\end{minipage}};\end{tikzpicture}\end{minipage}\par\vspace{2.4mm}\noindent\begin{minipage}[t]{\linewidth}\centering
\begin{tikzpicture}\node[draw=mgreenL,rounded corners=6pt,line width=.8pt,inner sep=6pt,text width=.94\linewidth] {%
\begin{minipage}{\linewidth}\raggedright
\textbf{\large Schema + command recovery}\hspace{0.6em plus 1fill}\pill{mgreenL}{LOCKED}\\[-1mm]
{\fontsize{7.2}{8.0}\selectfont\color{mgrayL}LOCKED WINNER}
\vspace{1.5mm}
\centering\resizebox{\linewidth}{!}{\begin{tikzpicture}[node distance=2mm]
\node[lat,minimum width=14mm] (x) {Sensors\\\textbf{Channel $p$=0.15}};
\node[enc,minimum width=16mm,right=2mm of x] (e) {Encoder\\$d=256$, $L=6$};
\node[prd,minimum width=16mm,right=2mm of e] (p) {Predictor\\\textbf{Token}};
\node[lat,minimum width=9mm,right=2mm of p] (zh) {$\hat z^{+}$};
\node[act,minimum width=14mm,below=3mm of p] (ac) {future commands};
\node[tgt,minimum width=16mm,below=3mm of e] (te) {EMA target\\0.996 fixed};
\draw[ar] (x)--(e); \draw[ar] (e)--(p); \draw[ar] (p)--(zh); \draw[ar] (ac)--(p); \draw[ema] (e)--(te);
\end{tikzpicture}}
\raggedright\vspace{1mm}
\pill{mblueL}{mask: Channel $p$=0.15} \pill{myellL}{command: Token} \pill{mgreenL}{VICReg: Per-horizon} \pill{mgrayL}{place: ctx+tgt}\\[1mm]
\pill{mgrayL}{$\lambda_{lat}$=1.0} \pill{mgreenL}{$\lambda_{sch}$=0.1} \pill{myellL}{$\lambda_{act}$=0.05} \pill{mblueL}{target: Per-horizon} \pill{mgrayL}{latent LN: off}\\[1mm]
\pill{mpinkL}{EMA: 0.996 fixed} \pill{mblueL}{norm: z-score} \pill{mgrayL}{head: Fresh}
\vspace{1.2mm}
{\fontsize{6.9}{7.6}\selectfont \textbf{Source validation:} Clean 0.8216 $\pm$ 0.0089 | schema 0.9183 | cmd 1.0583 | eff.rank 0.4273\\\textbf{Selection:} Discovery \#5 (-0.296); confirmation \#1 (-0.830).}
\end{minipage}};\end{tikzpicture}\end{minipage}\par\vspace{2.4mm}\noindent\begin{minipage}[t]{\linewidth}\centering
\begin{tikzpicture}\node[draw=mblueL,rounded corners=6pt,line width=.8pt,inner sep=6pt,text width=.94\linewidth] {%
\begin{minipage}{\linewidth}\raggedright
\textbf{\large FiLM command injection}\hspace{0.6em plus 1fill}\pill{mblueL}{5/5 seeds}\\[-1mm]
{\fontsize{7.2}{8.0}\selectfont\color{mgrayL}VALID}
\vspace{1.5mm}
\centering\resizebox{\linewidth}{!}{\begin{tikzpicture}[node distance=2mm]
\node[lat,minimum width=14mm] (x) {Sensors\\\textbf{Mixed 35\%}};
\node[enc,minimum width=16mm,right=2mm of x] (e) {Encoder\\$d=256$, $L=6$};
\node[prd,minimum width=16mm,right=2mm of e] (p) {Predictor\\\textbf{FiLM}};
\node[lat,minimum width=9mm,right=2mm of p] (zh) {$\hat z^{+}$};
\node[act,minimum width=14mm,below=3mm of p] (ac) {future commands};
\node[tgt,minimum width=16mm,below=3mm of e] (te) {EMA target\\0.996 fixed};
\draw[ar] (x)--(e); \draw[ar] (e)--(p); \draw[ar] (p)--(zh); \draw[ar] (ac)--(p); \draw[ema] (e)--(te);
\end{tikzpicture}}
\raggedright\vspace{1mm}
\pill{mblueL}{mask: Mixed 0.35} \pill{myellL}{command: FiLM} \pill{mgreenL}{VICReg: Per-horizon} \pill{mgrayL}{place: ctx+tgt}\\[1mm]
\pill{mgrayL}{$\lambda_{lat}$=1.0} \pill{mgreenL}{$\lambda_{sch}$=0.1} \pill{myellL}{$\lambda_{act}$=0.05} \pill{mblueL}{target: Per-horizon} \pill{mgrayL}{latent LN: off}\\[1mm]
\pill{mpinkL}{EMA: 0.996 fixed} \pill{mblueL}{norm: z-score} \pill{mgrayL}{head: Fresh}
\vspace{1.2mm}
{\fontsize{6.9}{7.6}\selectfont \textbf{Source validation:} Clean 0.8346 $\pm$ 0.0067 | schema 0.952 | cmd 1.0346 | eff.rank 0.3289\\\textbf{Selection:} Discovery \#3 (-0.318); confirmation \#4 (0.149).}
\end{minipage}};\end{tikzpicture}\end{minipage}\par\vspace{2.4mm}\noindent\begin{minipage}[t]{\linewidth}\centering
\begin{tikzpicture}\node[draw=mblueL,rounded corners=6pt,line width=.8pt,inner sep=6pt,text width=.94\linewidth] {%
\begin{minipage}{\linewidth}\raggedright
\textbf{\large Cross-attention command injection}\hspace{0.6em plus 1fill}\pill{mblueL}{5/5 seeds}\\[-1mm]
{\fontsize{7.2}{8.0}\selectfont\color{mgrayL}VALID}
\vspace{1.5mm}
\centering\resizebox{\linewidth}{!}{\begin{tikzpicture}[node distance=2mm]
\node[lat,minimum width=14mm] (x) {Sensors\\\textbf{Mixed 35\%}};
\node[enc,minimum width=16mm,right=2mm of x] (e) {Encoder\\$d=256$, $L=6$};
\node[prd,minimum width=16mm,right=2mm of e] (p) {Predictor\\\textbf{Cross-attn}};
\node[lat,minimum width=9mm,right=2mm of p] (zh) {$\hat z^{+}$};
\node[act,minimum width=14mm,below=3mm of p] (ac) {future commands};
\node[tgt,minimum width=16mm,below=3mm of e] (te) {EMA target\\0.996 fixed};
\draw[ar] (x)--(e); \draw[ar] (e)--(p); \draw[ar] (p)--(zh); \draw[ar] (ac)--(p); \draw[ema] (e)--(te);
\end{tikzpicture}}
\raggedright\vspace{1mm}
\pill{mblueL}{mask: Mixed 0.35} \pill{myellL}{command: Cross-attn} \pill{mgreenL}{VICReg: Per-horizon} \pill{mgrayL}{place: ctx+tgt}\\[1mm]
\pill{mgrayL}{$\lambda_{lat}$=1.0} \pill{mgreenL}{$\lambda_{sch}$=0.1} \pill{myellL}{$\lambda_{act}$=0.05} \pill{mblueL}{target: Per-horizon} \pill{mgrayL}{latent LN: off}\\[1mm]
\pill{mpinkL}{EMA: 0.996 fixed} \pill{mblueL}{norm: z-score} \pill{mgrayL}{head: Fresh}
\vspace{1.2mm}
{\fontsize{6.9}{7.6}\selectfont \textbf{Source validation:} Clean 0.8333 $\pm$ 0.0034 | schema 0.9488 | cmd 1.0336 | eff.rank 0.3149\\\textbf{Selection:} Discovery \#4 (-0.3); confirmation \#3 (0.142).}
\end{minipage}};\end{tikzpicture}\end{minipage}\par\vspace{2.4mm}\noindent\begin{minipage}[t]{\linewidth}\centering
\begin{tikzpicture}\node[draw=mblueL,rounded corners=6pt,line width=.8pt,inner sep=6pt,text width=.94\linewidth] {%
\begin{minipage}{\linewidth}\raggedright
\textbf{\large Event masking}\hspace{0.6em plus 1fill}\pill{mblueL}{4/4 seeds}\\[-1mm]
{\fontsize{7.2}{8.0}\selectfont\color{mgrayL}VALID}
\vspace{1.5mm}
\centering\resizebox{\linewidth}{!}{\begin{tikzpicture}[node distance=2mm]
\node[lat,minimum width=14mm] (x) {Sensors\\\textbf{Event 35\%}};
\node[enc,minimum width=16mm,right=2mm of x] (e) {Encoder\\$d=256$, $L=6$};
\node[prd,minimum width=16mm,right=2mm of e] (p) {Predictor\\\textbf{Cross-attn}};
\node[lat,minimum width=9mm,right=2mm of p] (zh) {$\hat z^{+}$};
\node[act,minimum width=14mm,below=3mm of p] (ac) {future commands};
\node[tgt,minimum width=16mm,below=3mm of e] (te) {EMA target\\0.996 fixed};
\draw[ar] (x)--(e); \draw[ar] (e)--(p); \draw[ar] (p)--(zh); \draw[ar] (ac)--(p); \draw[ema] (e)--(te);
\end{tikzpicture}}
\raggedright\vspace{1mm}
\pill{mblueL}{mask: Event 0.35} \pill{myellL}{command: Cross-attn} \pill{mgreenL}{VICReg: Per-horizon} \pill{mgrayL}{place: ctx+tgt}\\[1mm]
\pill{mgrayL}{$\lambda_{lat}$=1.0} \pill{mgreenL}{$\lambda_{sch}$=0.1} \pill{myellL}{$\lambda_{act}$=0.05} \pill{mblueL}{target: Per-horizon} \pill{mgrayL}{latent LN: off}\\[1mm]
\pill{mpinkL}{EMA: 0.996 fixed} \pill{mblueL}{norm: z-score} \pill{mgrayL}{head: Fresh}
\vspace{1.2mm}
{\fontsize{6.9}{7.6}\selectfont \textbf{Source validation:} Clean 0.8505 $\pm$ 0.0077 | schema 0.9567 | cmd 1.0295 | eff.rank 0.3278\\\textbf{Selection:} Discovery \#6 (-0.291).}
\end{minipage}};\end{tikzpicture}\end{minipage}\par\vspace{2.4mm}\noindent\begin{minipage}[t]{\linewidth}\centering
\begin{tikzpicture}\node[draw=mblueL,rounded corners=6pt,line width=.8pt,inner sep=6pt,text width=.94\linewidth] {%
\begin{minipage}{\linewidth}\raggedright
\textbf{\large Heavy channel masking}\hspace{0.6em plus 1fill}\pill{mblueL}{4/4 seeds}\\[-1mm]
{\fontsize{7.2}{8.0}\selectfont\color{mgrayL}VALID}
\vspace{1.5mm}
\centering\resizebox{\linewidth}{!}{\begin{tikzpicture}[node distance=2mm]
\node[lat,minimum width=14mm] (x) {Sensors\\\textbf{Channel $p$=0.15}};
\node[enc,minimum width=16mm,right=2mm of x] (e) {Encoder\\$d=256$, $L=6$};
\node[prd,minimum width=16mm,right=2mm of e] (p) {Predictor\\\textbf{Cross-attn}};
\node[lat,minimum width=9mm,right=2mm of p] (zh) {$\hat z^{+}$};
\node[act,minimum width=14mm,below=3mm of p] (ac) {future commands};
\node[tgt,minimum width=16mm,below=3mm of e] (te) {EMA target\\0.996 fixed};
\draw[ar] (x)--(e); \draw[ar] (e)--(p); \draw[ar] (p)--(zh); \draw[ar] (ac)--(p); \draw[ema] (e)--(te);
\end{tikzpicture}}
\raggedright\vspace{1mm}
\pill{mblueL}{mask: Channel $p$=0.15} \pill{myellL}{command: Cross-attn} \pill{mgreenL}{VICReg: Per-horizon} \pill{mgrayL}{place: ctx+tgt}\\[1mm]
\pill{mgrayL}{$\lambda_{lat}$=1.0} \pill{mgreenL}{$\lambda_{sch}$=0.2} \pill{myellL}{$\lambda_{act}$=0.05} \pill{mblueL}{target: Per-horizon} \pill{mgrayL}{latent LN: off}\\[1mm]
\pill{mpinkL}{EMA: 0.996 fixed} \pill{mblueL}{norm: z-score} \pill{mgrayL}{head: Fresh}
\vspace{1.2mm}
{\fontsize{6.9}{7.6}\selectfont \textbf{Source validation:} Clean 0.8213 $\pm$ 0.0075 | schema 0.9287 | cmd 1.0549 | eff.rank 0.416\\\textbf{Selection:} Discovery \#13 (-0.232).}
\end{minipage}};\end{tikzpicture}\end{minipage}\par\vspace{2.4mm}\noindent\begin{minipage}[t]{\linewidth}\centering
\begin{tikzpicture}\node[draw=mblueL,rounded corners=6pt,line width=.8pt,inner sep=6pt,text width=.94\linewidth] {%
\begin{minipage}{\linewidth}\raggedright
\textbf{\large Light mixed masking}\hspace{0.6em plus 1fill}\pill{mblueL}{4/4 seeds}\\[-1mm]
{\fontsize{7.2}{8.0}\selectfont\color{mgrayL}VALID}
\vspace{1.5mm}
\centering\resizebox{\linewidth}{!}{\begin{tikzpicture}[node distance=2mm]
\node[lat,minimum width=14mm] (x) {Sensors\\\textbf{Mixed 20\%}};
\node[enc,minimum width=16mm,right=2mm of x] (e) {Encoder\\$d=256$, $L=6$};
\node[prd,minimum width=16mm,right=2mm of e] (p) {Predictor\\\textbf{Cross-attn}};
\node[lat,minimum width=9mm,right=2mm of p] (zh) {$\hat z^{+}$};
\node[act,minimum width=14mm,below=3mm of p] (ac) {future commands};
\node[tgt,minimum width=16mm,below=3mm of e] (te) {EMA target\\0.996 fixed};
\draw[ar] (x)--(e); \draw[ar] (e)--(p); \draw[ar] (p)--(zh); \draw[ar] (ac)--(p); \draw[ema] (e)--(te);
\end{tikzpicture}}
\raggedright\vspace{1mm}
\pill{mblueL}{mask: Mixed 0.2} \pill{myellL}{command: Cross-attn} \pill{mgreenL}{VICReg: Per-horizon} \pill{mgrayL}{place: ctx+tgt}\\[1mm]
\pill{mgrayL}{$\lambda_{lat}$=1.0} \pill{mgreenL}{$\lambda_{sch}$=0.1} \pill{myellL}{$\lambda_{act}$=0.05} \pill{mblueL}{target: Per-horizon} \pill{mgrayL}{latent LN: off}\\[1mm]
\pill{mpinkL}{EMA: 0.996 fixed} \pill{mblueL}{norm: z-score} \pill{mgrayL}{head: Fresh}
\vspace{1.2mm}
{\fontsize{6.9}{7.6}\selectfont \textbf{Source validation:} Clean 0.8269 $\pm$ 0.0102 | schema 0.9432 | cmd 1.047 | eff.rank 0.3687\\\textbf{Selection:} Discovery \#9 (-0.28).}
\end{minipage}};\end{tikzpicture}\end{minipage}\par\vspace{2.4mm}\noindent\begin{minipage}[t]{\linewidth}\centering
\begin{tikzpicture}\node[draw=mblueL,rounded corners=6pt,line width=.8pt,inner sep=6pt,text width=.94\linewidth] {%
\begin{minipage}{\linewidth}\raggedright
\textbf{\large Latent LayerNorm ON}\hspace{0.6em plus 1fill}\pill{mblueL}{4/4 seeds}\\[-1mm]
{\fontsize{7.2}{8.0}\selectfont\color{mgrayL}VALID}
\vspace{1.5mm}
\centering\resizebox{\linewidth}{!}{\begin{tikzpicture}[node distance=2mm]
\node[lat,minimum width=14mm] (x) {Sensors\\\textbf{Mixed 35\%}};
\node[enc,minimum width=16mm,right=2mm of x] (e) {Encoder\\$d=256$, $L=6$};
\node[prd,minimum width=16mm,right=2mm of e] (p) {Predictor\\\textbf{Cross-attn}};
\node[lat,minimum width=9mm,right=2mm of p] (zh) {$\hat z^{+}$};
\node[act,minimum width=14mm,below=3mm of p] (ac) {future commands};
\node[tgt,minimum width=16mm,below=3mm of e] (te) {EMA target\\0.996 fixed};
\draw[ar] (x)--(e); \draw[ar] (e)--(p); \draw[ar] (p)--(zh); \draw[ar] (ac)--(p); \draw[ema] (e)--(te);
\end{tikzpicture}}
\raggedright\vspace{1mm}
\pill{mblueL}{mask: Mixed 0.35} \pill{myellL}{command: Cross-attn} \pill{mgreenL}{VICReg: Per-horizon} \pill{mgrayL}{place: ctx+tgt}\\[1mm]
\pill{mgrayL}{$\lambda_{lat}$=1.0} \pill{mgreenL}{$\lambda_{sch}$=0.1} \pill{myellL}{$\lambda_{act}$=0.05} \pill{mblueL}{target: Per-horizon} \pill{mgrayL}{latent LN: ON}\\[1mm]
\pill{mpinkL}{EMA: 0.996 fixed} \pill{mblueL}{norm: z-score} \pill{mgrayL}{head: Fresh}
\vspace{1.2mm}
{\fontsize{6.9}{7.6}\selectfont \textbf{Source validation:} Clean 0.8143 $\pm$ 0.0164 | schema 0.9545 | cmd 1.074 | eff.rank 0.3616\\\textbf{Selection:} Discovery \#10 (-0.273).}
\end{minipage}};\end{tikzpicture}\end{minipage}\par\vspace{2.4mm}\noindent\begin{minipage}[t]{\linewidth}\centering
\begin{tikzpicture}\node[draw=mblueL,rounded corners=6pt,line width=.8pt,inner sep=6pt,text width=.94\linewidth] {%
\begin{minipage}{\linewidth}\raggedright
\textbf{\large Pooled VICReg/target}\hspace{0.6em plus 1fill}\pill{mblueL}{4/4 seeds}\\[-1mm]
{\fontsize{7.2}{8.0}\selectfont\color{mgrayL}VALID}
\vspace{1.5mm}
\centering\resizebox{\linewidth}{!}{\begin{tikzpicture}[node distance=2mm]
\node[lat,minimum width=14mm] (x) {Sensors\\\textbf{Mixed 35\%}};
\node[enc,minimum width=16mm,right=2mm of x] (e) {Encoder\\$d=256$, $L=6$};
\node[prd,minimum width=16mm,right=2mm of e] (p) {Predictor\\\textbf{Cross-attn}};
\node[lat,minimum width=9mm,right=2mm of p] (zh) {$\hat z^{+}$};
\node[act,minimum width=14mm,below=3mm of p] (ac) {future commands};
\node[tgt,minimum width=16mm,below=3mm of e] (te) {EMA target\\0.996 fixed};
\draw[ar] (x)--(e); \draw[ar] (e)--(p); \draw[ar] (p)--(zh); \draw[ar] (ac)--(p); \draw[ema] (e)--(te);
\end{tikzpicture}}
\raggedright\vspace{1mm}
\pill{mblueL}{mask: Mixed 0.35} \pill{myellL}{command: Cross-attn} \pill{mgreenL}{VICReg: Pooled} \pill{mgrayL}{place: ctx+tgt}\\[1mm]
\pill{mgrayL}{$\lambda_{lat}$=1.0} \pill{mgreenL}{$\lambda_{sch}$=0.1} \pill{myellL}{$\lambda_{act}$=0.05} \pill{mblueL}{target: Pooled} \pill{mgrayL}{latent LN: off}\\[1mm]
\pill{mpinkL}{EMA: 0.996 fixed} \pill{mblueL}{norm: z-score} \pill{mgrayL}{head: Fresh}
\vspace{1.2mm}
{\fontsize{6.9}{7.6}\selectfont \textbf{Source validation:} Clean 0.8139 $\pm$ 0.0117 | schema 0.9228 | cmd 1.079 | eff.rank 0.4005\\\textbf{Selection:} Discovery \#14 (-0.229).}
\end{minipage}};\end{tikzpicture}\end{minipage}\par\vspace{2.4mm}\noindent\begin{minipage}[t]{\linewidth}\centering
\begin{tikzpicture}\node[draw=mblueL,rounded corners=6pt,line width=.8pt,inner sep=6pt,text width=.94\linewidth] {%
\begin{minipage}{\linewidth}\raggedright
\textbf{\large Low latent weight}\hspace{0.6em plus 1fill}\pill{mblueL}{4/4 seeds}\\[-1mm]
{\fontsize{7.2}{8.0}\selectfont\color{mgrayL}VALID}
\vspace{1.5mm}
\centering\resizebox{\linewidth}{!}{\begin{tikzpicture}[node distance=2mm]
\node[lat,minimum width=14mm] (x) {Sensors\\\textbf{Mixed 35\%}};
\node[enc,minimum width=16mm,right=2mm of x] (e) {Encoder\\$d=256$, $L=6$};
\node[prd,minimum width=16mm,right=2mm of e] (p) {Predictor\\\textbf{Cross-attn}};
\node[lat,minimum width=9mm,right=2mm of p] (zh) {$\hat z^{+}$};
\node[act,minimum width=14mm,below=3mm of p] (ac) {future commands};
\node[tgt,minimum width=16mm,below=3mm of e] (te) {EMA target\\0.996 fixed};
\draw[ar] (x)--(e); \draw[ar] (e)--(p); \draw[ar] (p)--(zh); \draw[ar] (ac)--(p); \draw[ema] (e)--(te);
\end{tikzpicture}}
\raggedright\vspace{1mm}
\pill{mblueL}{mask: Mixed 0.35} \pill{myellL}{command: Cross-attn} \pill{mgreenL}{VICReg: Per-horizon} \pill{mgrayL}{place: ctx+tgt}\\[1mm]
\pill{mgrayL}{$\lambda_{lat}$=0.5} \pill{mgreenL}{$\lambda_{sch}$=0.1} \pill{myellL}{$\lambda_{act}$=0.05} \pill{mblueL}{target: Per-horizon} \pill{mgrayL}{latent LN: off}\\[1mm]
\pill{mpinkL}{EMA: 0.996 fixed} \pill{mblueL}{norm: z-score} \pill{mgrayL}{head: Fresh}
\vspace{1.2mm}
{\fontsize{6.9}{7.6}\selectfont \textbf{Source validation:} Clean 0.8019 $\pm$ 0.0088 | schema 0.9065 | cmd 1.0925 | eff.rank 0.4145\\\textbf{Selection:} Discovery \#12 (-0.237).}
\end{minipage}};\end{tikzpicture}\end{minipage}\par\vspace{2.4mm}\noindent\begin{minipage}[t]{\linewidth}\centering
\begin{tikzpicture}\node[draw=mpinkL,rounded corners=6pt,line width=.8pt,inner sep=6pt,text width=.94\linewidth] {%
\begin{minipage}{\linewidth}\raggedright
\textbf{\large High latent weight}\hspace{0.6em plus 1fill}\pill{mpinkL}{1 COLLAPSE}\\[-1mm]
{\fontsize{7.2}{8.0}\selectfont\color{mgrayL}PARTIAL: 1 COLLAPSED SEED}
\vspace{1.5mm}
\centering\resizebox{\linewidth}{!}{\begin{tikzpicture}[node distance=2mm]
\node[lat,minimum width=14mm] (x) {Sensors\\\textbf{Mixed 35\%}};
\node[enc,minimum width=16mm,right=2mm of x] (e) {Encoder\\$d=256$, $L=6$};
\node[prd,minimum width=16mm,right=2mm of e] (p) {Predictor\\\textbf{Cross-attn}};
\node[lat,minimum width=9mm,right=2mm of p] (zh) {$\hat z^{+}$};
\node[act,minimum width=14mm,below=3mm of p] (ac) {future commands};
\node[tgt,minimum width=16mm,below=3mm of e] (te) {EMA target\\0.996 fixed};
\draw[ar] (x)--(e); \draw[ar] (e)--(p); \draw[ar] (p)--(zh); \draw[ar] (ac)--(p); \draw[ema] (e)--(te);
\end{tikzpicture}}
\raggedright\vspace{1mm}
\pill{mblueL}{mask: Mixed 0.35} \pill{myellL}{command: Cross-attn} \pill{mgreenL}{VICReg: Per-horizon} \pill{mgrayL}{place: ctx+tgt}\\[1mm]
\pill{mgrayL}{$\lambda_{lat}$=2.0} \pill{mgreenL}{$\lambda_{sch}$=0.1} \pill{myellL}{$\lambda_{act}$=0.05} \pill{mblueL}{target: Per-horizon} \pill{mgrayL}{latent LN: off}\\[1mm]
\pill{mpinkL}{EMA: 0.996 fixed} \pill{mblueL}{norm: z-score} \pill{mgrayL}{head: Fresh}
\vspace{1.2mm}
{\fontsize{6.9}{7.6}\selectfont \textbf{Source validation:} Clean 0.8438 $\pm$ 0.0015 | schema 0.9499 | cmd 1.0214 | eff.rank 0.2906\\\textbf{Selection:} Discovery \#2 (-0.323); confirmation \#5 (1.177); invalid: ssl\_latent\_collapse.}
\end{minipage}};\end{tikzpicture}\end{minipage}\par\vspace{2.4mm}\noindent\begin{minipage}[t]{\linewidth}\centering
\begin{tikzpicture}\node[draw=mblueL,rounded corners=6pt,line width=.8pt,inner sep=6pt,text width=.94\linewidth] {%
\begin{minipage}{\linewidth}\raggedright
\textbf{\large Scheduled EMA}\hspace{0.6em plus 1fill}\pill{mblueL}{4/4 seeds}\\[-1mm]
{\fontsize{7.2}{8.0}\selectfont\color{mgrayL}VALID}
\vspace{1.5mm}
\centering\resizebox{\linewidth}{!}{\begin{tikzpicture}[node distance=2mm]
\node[lat,minimum width=14mm] (x) {Sensors\\\textbf{Mixed 35\%}};
\node[enc,minimum width=16mm,right=2mm of x] (e) {Encoder\\$d=256$, $L=6$};
\node[prd,minimum width=16mm,right=2mm of e] (p) {Predictor\\\textbf{Cross-attn}};
\node[lat,minimum width=9mm,right=2mm of p] (zh) {$\hat z^{+}$};
\node[act,minimum width=14mm,below=3mm of p] (ac) {future commands};
\node[tgt,minimum width=16mm,below=3mm of e] (te) {EMA target\\0.99$\to$0.9999};
\draw[ar] (x)--(e); \draw[ar] (e)--(p); \draw[ar] (p)--(zh); \draw[ar] (ac)--(p); \draw[ema] (e)--(te);
\end{tikzpicture}}
\raggedright\vspace{1mm}
\pill{mblueL}{mask: Mixed 0.35} \pill{myellL}{command: Cross-attn} \pill{mgreenL}{VICReg: Per-horizon} \pill{mgrayL}{place: ctx+tgt}\\[1mm]
\pill{mgrayL}{$\lambda_{lat}$=1.0} \pill{mgreenL}{$\lambda_{sch}$=0.1} \pill{myellL}{$\lambda_{act}$=0.05} \pill{mblueL}{target: Per-horizon} \pill{mgrayL}{latent LN: off}\\[1mm]
\pill{mpinkL}{EMA: 0.99$\to$0.9999} \pill{mblueL}{norm: z-score} \pill{mgrayL}{head: Fresh}
\vspace{1.2mm}
{\fontsize{6.9}{7.6}\selectfont \textbf{Source validation:} Clean 0.8345 $\pm$ 0.003 | schema 0.9574 | cmd 1.0311 | eff.rank 0.3002\\\textbf{Selection:} Discovery \#7 (-0.291).}
\end{minipage}};\end{tikzpicture}\end{minipage}\par\vspace{2.4mm}\noindent\begin{minipage}[t]{\linewidth}\centering
\begin{tikzpicture}\node[draw=mredL,rounded corners=6pt,line width=.8pt,inner sep=6pt,text width=.94\linewidth] {%
\begin{minipage}{\linewidth}\raggedright
\textbf{\large Robust source normalization}\hspace{0.6em plus 1fill}\pill{mredL}{IQR PATHOLOGY}\\[-1mm]
{\fontsize{7.2}{8.0}\selectfont\color{mgrayL}PATHOLOGICAL ROBUST NORMALIZATION}
\vspace{1.5mm}
\centering\resizebox{\linewidth}{!}{\begin{tikzpicture}[node distance=2mm]
\node[lat,minimum width=14mm] (x) {Sensors\\\textbf{Mixed 35\%}};
\node[enc,minimum width=16mm,right=2mm of x] (e) {Encoder\\$d=256$, $L=6$};
\node[prd,minimum width=16mm,right=2mm of e] (p) {Predictor\\\textbf{Cross-attn}};
\node[lat,minimum width=9mm,right=2mm of p] (zh) {$\hat z^{+}$};
\node[act,minimum width=14mm,below=3mm of p] (ac) {future commands};
\node[tgt,minimum width=16mm,below=3mm of e] (te) {EMA target\\0.99$\to$0.9999};
\draw[ar] (x)--(e); \draw[ar] (e)--(p); \draw[ar] (p)--(zh); \draw[ar] (ac)--(p); \draw[ema] (e)--(te);
\end{tikzpicture}}
\raggedright\vspace{1mm}
\pill{mblueL}{mask: Mixed 0.35} \pill{myellL}{command: Cross-attn} \pill{mgreenL}{VICReg: Per-horizon} \pill{mgrayL}{place: ctx+tgt}\\[1mm]
\pill{mgrayL}{$\lambda_{lat}$=1.0} \pill{mgreenL}{$\lambda_{sch}$=0.1} \pill{myellL}{$\lambda_{act}$=0.05} \pill{mblueL}{target: Per-horizon} \pill{mgrayL}{latent LN: off}\\[1mm]
\pill{mpinkL}{EMA: 0.99$\to$0.9999} \pill{mblueL}{norm: Robust} \pill{mgrayL}{head: Fresh}
\vspace{1.2mm}
{\fontsize{6.9}{7.6}\selectfont \textbf{Source validation:} Clean 0.9813 $\pm$ 0.0360 | schema 0.9814 | cmd 1.0088 | eff.rank 0.0213\\\textbf{Selection:} Discovery \#17 (6.25).}
\end{minipage}};\end{tikzpicture}\end{minipage}\par\vspace{2.4mm}\noindent\begin{minipage}[t]{\linewidth}\centering
\begin{tikzpicture}\node[draw=mblueL,rounded corners=6pt,line width=.8pt,inner sep=6pt,text width=.94\linewidth] {%
\begin{minipage}{\linewidth}\raggedright
\textbf{\large Compact model}\hspace{0.6em plus 1fill}\pill{mblueL}{4/4 seeds}\\[-1mm]
{\fontsize{7.2}{8.0}\selectfont\color{mgrayL}VALID}
\vspace{1.5mm}
\centering\resizebox{\linewidth}{!}{\begin{tikzpicture}[node distance=2mm]
\node[lat,minimum width=14mm] (x) {Sensors\\\textbf{Mixed 35\%}};
\node[enc,minimum width=16mm,right=2mm of x] (e) {Encoder\\$d=128$, $L=4$};
\node[prd,minimum width=16mm,right=2mm of e] (p) {Predictor\\\textbf{Cross-attn}};
\node[lat,minimum width=9mm,right=2mm of p] (zh) {$\hat z^{+}$};
\node[act,minimum width=14mm,below=3mm of p] (ac) {future commands};
\node[tgt,minimum width=16mm,below=3mm of e] (te) {EMA target\\0.99$\to$0.9999};
\draw[ar] (x)--(e); \draw[ar] (e)--(p); \draw[ar] (p)--(zh); \draw[ar] (ac)--(p); \draw[ema] (e)--(te);
\end{tikzpicture}}
\raggedright\vspace{1mm}
\pill{mblueL}{mask: Mixed 0.35} \pill{myellL}{command: Cross-attn} \pill{mgreenL}{VICReg: Per-horizon} \pill{mgrayL}{place: ctx+tgt}\\[1mm]
\pill{mgrayL}{$\lambda_{lat}$=1.0} \pill{mgreenL}{$\lambda_{sch}$=0.1} \pill{myellL}{$\lambda_{act}$=0.05} \pill{mblueL}{target: Per-horizon} \pill{mgrayL}{latent LN: off}\\[1mm]
\pill{mpinkL}{EMA: 0.99$\to$0.9999} \pill{mblueL}{norm: z-score} \pill{mgrayL}{head: Fresh}
\vspace{1.2mm}
{\fontsize{6.9}{7.6}\selectfont \textbf{Source validation:} Clean 0.8485 $\pm$ 0.0006 | schema 0.9439 | cmd 1.0119 | eff.rank 0.3485\\\textbf{Selection:} Discovery \#16 (0.462).}
\end{minipage}};\end{tikzpicture}\end{minipage}\par\vspace{2.4mm}\noindent\begin{minipage}[t]{\linewidth}\centering
\begin{tikzpicture}\node[draw=mblueL,rounded corners=6pt,line width=.8pt,inner sep=6pt,text width=.94\linewidth] {%
\begin{minipage}{\linewidth}\raggedright
\textbf{\large Medium-wide model}\hspace{0.6em plus 1fill}\pill{mblueL}{4/4 seeds}\\[-1mm]
{\fontsize{7.2}{8.0}\selectfont\color{mgrayL}VALID}
\vspace{1.5mm}
\centering\resizebox{\linewidth}{!}{\begin{tikzpicture}[node distance=2mm]
\node[lat,minimum width=14mm] (x) {Sensors\\\textbf{Mixed 35\%}};
\node[enc,minimum width=16mm,right=2mm of x] (e) {Encoder\\$d=384$, $L=6$};
\node[prd,minimum width=16mm,right=2mm of e] (p) {Predictor\\\textbf{Cross-attn}};
\node[lat,minimum width=9mm,right=2mm of p] (zh) {$\hat z^{+}$};
\node[act,minimum width=14mm,below=3mm of p] (ac) {future commands};
\node[tgt,minimum width=16mm,below=3mm of e] (te) {EMA target\\0.99$\to$0.9999};
\draw[ar] (x)--(e); \draw[ar] (e)--(p); \draw[ar] (p)--(zh); \draw[ar] (ac)--(p); \draw[ema] (e)--(te);
\end{tikzpicture}}
\raggedright\vspace{1mm}
\pill{mblueL}{mask: Mixed 0.35} \pill{myellL}{command: Cross-attn} \pill{mgreenL}{VICReg: Per-horizon} \pill{mgrayL}{place: ctx+tgt}\\[1mm]
\pill{mgrayL}{$\lambda_{lat}$=1.0} \pill{mgreenL}{$\lambda_{sch}$=0.1} \pill{myellL}{$\lambda_{act}$=0.05} \pill{mblueL}{target: Per-horizon} \pill{mgrayL}{latent LN: off}\\[1mm]
\pill{mpinkL}{EMA: 0.99$\to$0.9999} \pill{mblueL}{norm: z-score} \pill{mgrayL}{head: Fresh}
\vspace{1.2mm}
{\fontsize{6.9}{7.6}\selectfont \textbf{Source validation:} Clean 0.8148 $\pm$ 0.0016 | schema 0.9325 | cmd 1.0758 | eff.rank 0.305\\\textbf{Selection:} Discovery \#15 (-0.084).}
\end{minipage}};\end{tikzpicture}\end{minipage}\par\vspace{2.4mm}\noindent\begin{minipage}[t]{\linewidth}\centering
\begin{tikzpicture}\node[draw=mredL,rounded corners=6pt,line width=.8pt,inner sep=6pt,text width=.94\linewidth] {%
\begin{minipage}{\linewidth}\raggedright
\textbf{\large Large model}\hspace{0.6em plus 1fill}\pill{mredL}{COLLAPSE 3/4}\\[-1mm]
{\fontsize{7.2}{8.0}\selectfont\color{mgrayL}UNSTABLE: SSL LATENT COLLAPSE (3/4)}
\vspace{1.5mm}
\centering\resizebox{\linewidth}{!}{\begin{tikzpicture}[node distance=2mm]
\node[lat,minimum width=14mm] (x) {Sensors\\\textbf{Mixed 35\%}};
\node[enc,minimum width=16mm,right=2mm of x] (e) {Encoder\\$d=512$, $L=8$};
\node[prd,minimum width=16mm,right=2mm of e] (p) {Predictor\\\textbf{Cross-attn}};
\node[lat,minimum width=9mm,right=2mm of p] (zh) {$\hat z^{+}$};
\node[act,minimum width=14mm,below=3mm of p] (ac) {future commands};
\node[tgt,minimum width=16mm,below=3mm of e] (te) {EMA target\\0.99$\to$0.9999};
\draw[ar] (x)--(e); \draw[ar] (e)--(p); \draw[ar] (p)--(zh); \draw[ar] (ac)--(p); \draw[ema] (e)--(te);
\end{tikzpicture}}
\raggedright\vspace{1mm}
\pill{mblueL}{mask: Mixed 0.35} \pill{myellL}{command: Cross-attn} \pill{mgreenL}{VICReg: Per-horizon} \pill{mgrayL}{place: ctx+tgt}\\[1mm]
\pill{mgrayL}{$\lambda_{lat}$=1.0} \pill{mgreenL}{$\lambda_{sch}$=0.1} \pill{myellL}{$\lambda_{act}$=0.05} \pill{mblueL}{target: Per-horizon} \pill{mgrayL}{latent LN: off}\\[1mm]
\pill{mpinkL}{EMA: 0.99$\to$0.9999} \pill{mblueL}{norm: z-score} \pill{mgrayL}{head: Fresh}
\vspace{1.2mm}
{\fontsize{6.9}{7.6}\selectfont \textbf{Source validation:} Clean 0.7937 (1 seed) | schema 0.8857 | cmd 1.1222 | eff.rank 0.2462\\\textbf{Selection:} Discovery \#19 (-0.128); invalid: ssl\_latent\_collapse.}
\end{minipage}};\end{tikzpicture}\end{minipage}\par\vspace{2.4mm}\noindent\begin{minipage}[t]{\linewidth}\centering
\begin{tikzpicture}\node[draw=mblueL,rounded corners=6pt,line width=.8pt,inner sep=6pt,text width=.94\linewidth] {%
\begin{minipage}{\linewidth}\raggedright
\textbf{\large Head-contamination control}\hspace{0.6em plus 1fill}\pill{mblueL}{4/4 seeds}\\[-1mm]
{\fontsize{7.2}{8.0}\selectfont\color{mgrayL}VALID}
\vspace{1.5mm}
\centering\resizebox{\linewidth}{!}{\begin{tikzpicture}[node distance=2mm]
\node[lat,minimum width=14mm] (x) {Sensors\\\textbf{Mixed 35\%}};
\node[enc,minimum width=16mm,right=2mm of x] (e) {Encoder\\$d=256$, $L=6$};
\node[prd,minimum width=16mm,right=2mm of e] (p) {Predictor\\\textbf{Cross-attn}};
\node[lat,minimum width=9mm,right=2mm of p] (zh) {$\hat z^{+}$};
\node[act,minimum width=14mm,below=3mm of p] (ac) {future commands};
\node[tgt,minimum width=16mm,below=3mm of e] (te) {EMA target\\0.99$\to$0.9999};
\draw[ar] (x)--(e); \draw[ar] (e)--(p); \draw[ar] (p)--(zh); \draw[ar] (ac)--(p); \draw[ema] (e)--(te);
\end{tikzpicture}}
\raggedright\vspace{1mm}
\pill{mblueL}{mask: Mixed 0.35} \pill{myellL}{command: Cross-attn} \pill{mgreenL}{VICReg: Per-horizon} \pill{mgrayL}{place: ctx+tgt}\\[1mm]
\pill{mgrayL}{$\lambda_{lat}$=1.0} \pill{mgreenL}{$\lambda_{sch}$=0.1} \pill{myellL}{$\lambda_{act}$=0.05} \pill{mblueL}{target: Per-horizon} \pill{mgrayL}{latent LN: off}\\[1mm]
\pill{mpinkL}{EMA: 0.99$\to$0.9999} \pill{mblueL}{norm: z-score} \pill{mgrayL}{head: Loaded}
\vspace{1.2mm}
{\fontsize{6.9}{7.6}\selectfont \textbf{Source validation:} Clean 0.8359 $\pm$ 0.0036 | schema 0.9309 | cmd 1.0293 | eff.rank 0.2935\\\textbf{Selection:} Discovery \#8 (-0.288).}
\end{minipage}};\end{tikzpicture}\end{minipage}\par\vspace{2.4mm}\noindent\begin{minipage}[t]{\linewidth}\centering
\begin{tikzpicture}\node[draw=mredL,rounded corners=6pt,line width=.8pt,inner sep=6pt,text width=.94\linewidth] {%
\begin{minipage}{\linewidth}\raggedright
\textbf{\large Full candidate}\hspace{0.6em plus 1fill}\pill{mredL}{IQR PATHOLOGY}\\[-1mm]
{\fontsize{7.2}{8.0}\selectfont\color{mgrayL}PATHOLOGICAL ROBUST NORMALIZATION}
\vspace{1.5mm}
\centering\resizebox{\linewidth}{!}{\begin{tikzpicture}[node distance=2mm]
\node[lat,minimum width=14mm] (x) {Sensors\\\textbf{Mixed 35\%}};
\node[enc,minimum width=16mm,right=2mm of x] (e) {Encoder\\$d=384$, $L=6$};
\node[prd,minimum width=16mm,right=2mm of e] (p) {Predictor\\\textbf{Cross-attn}};
\node[lat,minimum width=9mm,right=2mm of p] (zh) {$\hat z^{+}$};
\node[act,minimum width=14mm,below=3mm of p] (ac) {future commands};
\node[tgt,minimum width=16mm,below=3mm of e] (te) {EMA target\\0.99$\to$0.9999};
\draw[ar] (x)--(e); \draw[ar] (e)--(p); \draw[ar] (p)--(zh); \draw[ar] (ac)--(p); \draw[ema] (e)--(te);
\end{tikzpicture}}
\raggedright\vspace{1mm}
\pill{mblueL}{mask: Mixed 0.35} \pill{myellL}{command: Cross-attn} \pill{mgreenL}{VICReg: Per-horizon} \pill{mgrayL}{place: ctx+tgt}\\[1mm]
\pill{mgrayL}{$\lambda_{lat}$=1.0} \pill{mgreenL}{$\lambda_{sch}$=0.2} \pill{myellL}{$\lambda_{act}$=0.1} \pill{mblueL}{target: Per-horizon} \pill{mgrayL}{latent LN: off}\\[1mm]
\pill{mpinkL}{EMA: 0.99$\to$0.9999} \pill{mblueL}{norm: Robust} \pill{mgrayL}{head: Fresh}
\vspace{1.2mm}
{\fontsize{6.9}{7.6}\selectfont \textbf{Source validation:} Clean 0.9557 $\pm$ 0.0013 | schema 0.9557 | cmd 1.0148 | eff.rank 0.011\\\textbf{Selection:} Discovery \#18 (6.3).}
\end{minipage}};\end{tikzpicture}\end{minipage}\par\vspace{2.4mm}
\section{Complete Candidate Configuration Matrix}
\footnotesize
Table~\ref{tab:all-configs} lists the exact design factors varied across the twenty candidates. For channel masking the value shown is the per-channel drop probability the code actually applies: the configuration's ratio field is unused in that mode, so the nominal settings 0.35 and 0.5 both ran at $p=0.15$, and the ``heavy channel masking'' candidate differs from the locked one only in command injection and schema weight. ``ctx+tgt'' denotes VICReg placement on context and EMA-target representations. All candidates use the same source-only data splits and evaluation protocol; only the listed architectural or training factors vary.
\begin{table*}[t]\centering\caption{Complete architecture-search configuration.}\label{tab:all-configs}\setlength{\tabcolsep}{5pt}\renewcommand{\arraystretch}{1.08}
\resizebox{\textwidth}{!}{%
\begin{tabular}{llcccccccccc}\toprule
\rowcolor{cvprblue!10}
Candidate & Mask & Cmd. & VICReg & $\lambda_{lat}$ & $\lambda_{sch}$ & $\lambda_{act}$ & EMA & $d/L$ & Norm & LN & FT head\\\midrule
Plain control & mixed/0.35 & Token & none & 1.0 & 0.0 & 0.0 & 0.996 & 256/6 & zscore & off & fresh\\
Per-horizon VICReg & mixed/0.35 & Token & per-h & 1.0 & 0.0 & 0.0 & 0.996 & 256/6 & zscore & off & fresh\\
Schema consistency & channel/$p$=0.15 & Token & per-h & 1.0 & 0.1 & 0.0 & 0.996 & 256/6 & zscore & off & fresh\\
Schema + command rec. & channel/$p$=0.15 & Token & per-h & 1.0 & 0.1 & 0.05 & 0.996 & 256/6 & zscore & off & fresh\\
FiLM command & mixed/0.35 & FiLM & per-h & 1.0 & 0.1 & 0.05 & 0.996 & 256/6 & zscore & off & fresh\\
Cross-attn command & mixed/0.35 & Cross-attn & per-h & 1.0 & 0.1 & 0.05 & 0.996 & 256/6 & zscore & off & fresh\\
Event masking & event/0.35 & Cross-attn & per-h & 1.0 & 0.1 & 0.05 & 0.996 & 256/6 & zscore & off & fresh\\
Heavy channel mask & channel/$p$=0.15 & Cross-attn & per-h & 1.0 & 0.2 & 0.05 & 0.996 & 256/6 & zscore & off & fresh\\
Light mixed mask & mixed/0.2 & Cross-attn & per-h & 1.0 & 0.1 & 0.05 & 0.996 & 256/6 & zscore & off & fresh\\
Latent LayerNorm & mixed/0.35 & Cross-attn & per-h & 1.0 & 0.1 & 0.05 & 0.996 & 256/6 & zscore & on & fresh\\
Pooled VICReg & mixed/0.35 & Cross-attn & pool & 1.0 & 0.1 & 0.05 & 0.996 & 256/6 & zscore & off & fresh\\
Low latent weight & mixed/0.35 & Cross-attn & per-h & 0.5 & 0.1 & 0.05 & 0.996 & 256/6 & zscore & off & fresh\\
High latent weight & mixed/0.35 & Cross-attn & per-h & 2.0 & 0.1 & 0.05 & 0.996 & 256/6 & zscore & off & fresh\\
Scheduled EMA & mixed/0.35 & Cross-attn & per-h & 1.0 & 0.1 & 0.05 & sched. & 256/6 & zscore & off & fresh\\
Robust normalization & mixed/0.35 & Cross-attn & per-h & 1.0 & 0.1 & 0.05 & sched. & 256/6 & robust & off & fresh\\
Compact model & mixed/0.35 & Cross-attn & per-h & 1.0 & 0.1 & 0.05 & sched. & 128/4 & zscore & off & fresh\\
Medium-wide model & mixed/0.35 & Cross-attn & per-h & 1.0 & 0.1 & 0.05 & sched. & 384/6 & zscore & off & fresh\\
Large model & mixed/0.35 & Cross-attn & per-h & 1.0 & 0.1 & 0.05 & sched. & 512/8 & zscore & off & fresh\\
Loaded-head control & mixed/0.35 & Cross-attn & per-h & 1.0 & 0.1 & 0.05 & sched. & 256/6 & zscore & off & loaded\\
Full candidate & mixed/0.35 & Cross-attn & per-h & 1.0 & 0.2 & 0.1 & sched. & 384/6 & robust & off & fresh\\
\bottomrule\end{tabular}}\end{table*}
\section{Source-Validation Search Outcomes}\label{app:outcomes}
\footnotesize
The table below reports the source-only architecture-search outcomes. Composite scores are z-normalized within their respective selection pools and therefore should not be compared across discovery and confirmation columns. The plain control was excluded because all four seeds collapsed. The large model collapsed in three of four discovery seeds. The two robust-normalization candidates exhibited the pathology associated with near-zero IQR channels. The final closure-stage stability check selected the schema-consistency variant with command recovery after seven valid seeds for the leading pair.
\begin{table*}[t]\centering\caption{Source-validation outcomes for the twenty candidates. Lower RMSE/schema-drop is better; command sensitivity greater than one indicates that shuffled future commands degrade prediction.}\setlength{\tabcolsep}{2.6pt}\renewcommand{\arraystretch}{1.12}\scriptsize\resizebox{\textwidth}{!}{%
\begin{tabular}{lcccccccccl}\toprule
\rowcolor{cvprblue!10}
Candidate & Seeds & Clean RMSE & Schema & Masked & Long & Calib. & Cmd. sens. & Eff. rank & Discovery / confirm.\\\midrule
Plain control & 0/4 & -- & -- & -- & -- & -- & -- & -- & excluded / excluded\\
Per-horizon VICReg & 7/7 & 0.8336 $\pm$ 0.0033 & 0.9494 & 0.8411 & 0.8797 & 0.1035 & 1.0350 & 0.3189 & \#1 (-0.346) / \#2 (0.093)\\
Schema consistency & 4/4 & 0.827 $\pm$ 0.0058 & 0.9252 & 0.8344 & 0.8747 & 0.1245 & 1.0437 & 0.3792 & \#11 (-0.272) / --\\
Schema + command rec. & 7/7 & 0.8216 $\pm$ 0.0089 & 0.9183 & 0.8298 & 0.8709 & 0.1315 & 1.0583 & 0.4273 & \#5 (-0.296) / \#1 (-0.830)\\
FiLM command & 5/5 & 0.8346 $\pm$ 0.0067 & 0.952 & 0.8412 & 0.8793 & 0.0957 & 1.0346 & 0.3289 & \#3 (-0.318) / \#4 (0.149)\\
Cross-attn command & 5/5 & 0.8333 $\pm$ 0.0034 & 0.9488 & 0.8415 & 0.8786 & 0.1003 & 1.0336 & 0.3149 & \#4 (-0.3) / \#3 (0.142)\\
Event masking & 4/4 & 0.8505 $\pm$ 0.0077 & 0.9567 & 0.8492 & 0.8976 & 0.125 & 1.0295 & 0.3278 & \#6 (-0.291) / --\\
Heavy channel mask & 4/4 & 0.8213 $\pm$ 0.0075 & 0.9287 & 0.8297 & 0.8713 & 0.1407 & 1.0549 & 0.416 & \#13 (-0.232) / --\\
Light mixed mask & 4/4 & 0.8269 $\pm$ 0.0102 & 0.9432 & 0.8364 & 0.8752 & 0.1171 & 1.047 & 0.3687 & \#9 (-0.28) / --\\
Latent LayerNorm & 4/4 & 0.8143 $\pm$ 0.0164 & 0.9545 & 0.8216 & 0.866 & 0.1268 & 1.074 & 0.3616 & \#10 (-0.273) / --\\
Pooled VICReg & 4/4 & 0.8139 $\pm$ 0.0117 & 0.9228 & 0.8221 & 0.8657 & 0.1435 & 1.079 & 0.4005 & \#14 (-0.229) / --\\
Low latent weight & 4/4 & 0.8019 $\pm$ 0.0088 & 0.9065 & 0.8127 & 0.8535 & 0.1421 & 1.0925 & 0.4145 & \#12 (-0.237) / --\\
High latent weight & 4/5 & 0.8438 $\pm$ 0.0015 & 0.9499 & 0.8521 & 0.8865 & 0.0976 & 1.0214 & 0.2906 & \#2 (-0.323) / \#5 (1.177)\\
Scheduled EMA & 4/4 & 0.8345 $\pm$ 0.003 & 0.9574 & 0.842 & 0.8804 & 0.0807 & 1.0311 & 0.3002 & \#7 (-0.291) / --\\
Robust normalization & 4/4 & 0.9813 $\pm$ 0.0360 & 0.9814 & 0.9816 & 1.0126 & 0.0399 & 1.0088 & 0.0213 & \#17 (6.25) / --\\
Compact model & 4/4 & 0.8485 $\pm$ 0.0006 & 0.9439 & 0.856 & 0.8883 & 0.0666 & 1.0119 & 0.3485 & \#16 (0.462) / --\\
Medium-wide model & 4/4 & 0.8148 $\pm$ 0.0016 & 0.9325 & 0.8237 & 0.8686 & 0.1147 & 1.0758 & 0.305 & \#15 (-0.084) / --\\
Large model & 1/4 & 0.7937 (1 seed) & 0.8857 & 0.8018 & 0.8558 & 0.1156 & 1.1222 & 0.2462 & \#19 (-0.128) / --\\
Loaded-head control & 4/4 & 0.8359 $\pm$ 0.0036 & 0.9309 & 0.8416 & 0.8818 & 0.0818 & 1.0293 & 0.2935 & \#8 (-0.288) / --\\
Full candidate & 4/4 & 0.9557 $\pm$ 0.0013 & 0.9557 & 0.9558 & 0.9866 & 0.0075 & 1.0148 & 0.011 & \#18 (6.3) / --\\
\bottomrule\end{tabular}}\end{table*}
\section{Architecture-Search Interpretation}
\small
\textbf{Anti-collapse is necessary.} The plain control removes VICReg and collapses for all four discovery seeds. This matches the corrected pretraining audit, in which target-encoder effective rank increased from approximately 5\% to 58\% after anti-collapse regularization was moved to the appropriate latent representations.

\textbf{Capacity scaling is not monotonic.} The large model ($d=512$, eight layers) collapses in three of four seeds, whereas the locked candidate uses $d=256$ and six layers. The search therefore does not support a ``larger is better'' interpretation at the available source-data scale.

\textbf{Robust normalization requires channel-specific safeguards.} The two robust-normalization candidates produce pathological errors because near-constant source channels can have an interquartile range close to zero. These candidates are retained in the appendix because they document a real failure mode of the search space rather than an omitted negative result.

\textbf{The locked candidate reflects the intended inductive bias.} The locked candidate combines channel masking, latent schema consistency, explicit command recovery, per-horizon VICReg placed on context and target, whose gradient reaches only the pooled context representation (the target half is monitored only), and a fresh downstream head. Its source-validation advantage is not defined by the lowest clean RMSE alone; the selection score also rewards schema-drop robustness, masked robustness, long-horizon performance, calibration, SSL validation, and command use.

\section{Official Forecasting Baseline Protocol}
\small
The official PatchTST (\texttt{yuqinie98/PatchTST}, commit \texttt{204c21e}, supervised tree) and iTransformer (\texttt{thuml/iTransformer}, commit \texttt{c2426e6}) repositories were run through a shared dataset adapter so that both consume the same session-bounded windows as our model: context 32 steps, prediction length 16, scored at horizons $\{1,2,4,8,16\}$, 1\,Hz grid, windows never crossing a session boundary (18,267 / 3,423 / 5,189 source windows and 2,457 target windows, identical to the counts of our model). Inputs are the 17 canonical sensors plus the 4 command channels (21 channels, all forecast jointly); RMSE is reported on the 17 sensors for the source test split and on the 10 channels shared with the target for the sealed target pass. Values are z-scored with the source-train normalizer of the main pipeline; the 7 sensors absent on the target are zero-imputed in normalized space and excluded from scoring. Both baselines were trained once (single seed 20260820, 100 epochs, early-stopping patience 15, batch 128, MSE loss, RevIN enabled without affine parameters, other hyperparameters at repository defaults) with no hyperparameter search; the same RevIN settings are reused for the RevIN variant of our model (\hyperref[app:revin]{App.~H}). The official SimMTM repository (commit \texttt{169513b}) diverged to NaN at two learning rates under the same protocol and is therefore not reported. Persistence on the same target windows gives RMSE 0.654; on the source test split it gives 1.128 (17 sensors). The official baselines forecast the command channels poorly (target RMSE 2.01 and 2.27), which is expected because commands are exogenous inputs rather than process outcomes.

\section{Concept of Study and V2 Roadmap}\label{app:concept}
\begin{figure*}[t]
  \centering
  \includegraphics[width=0.96\textwidth]{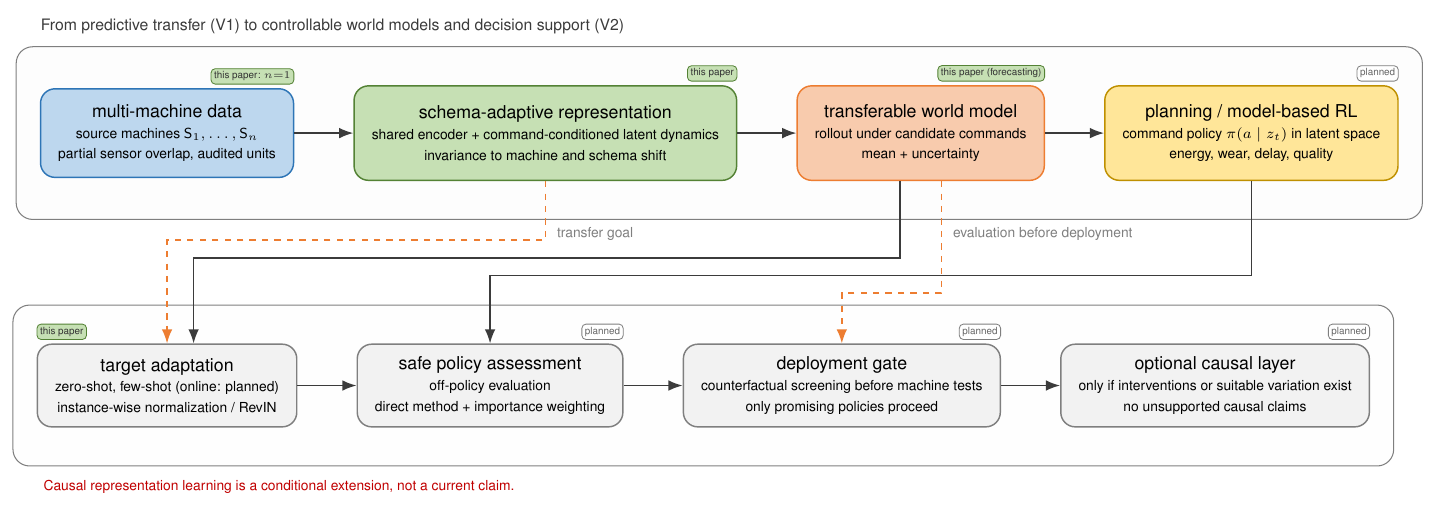}
  \caption{V1 establishes audited cross-machine predictive transfer with schema-adaptive latent dynamics. Green tags mark what this paper does (one source and one target machine; zero-shot and few-shot adaptation), grey tags what is planned. V2 extends this backbone to multi-machine representation learning, target adaptation with stronger normalization, command-conditioned planning or model-based RL, and policy assessment through off-policy evaluation before any physical deployment. Optional causal representation learning is shown as a conditional branch rather than as a current contribution.}
  \label{fig:v2}
\end{figure*}

\small
The overall concept of the study is a single pipeline from historical machine data to a deployable control decision:
\begin{enumerate}[leftmargin=*,nosep]
  \item \textbf{Source machine}: real production data (sensors and commands) from the source CNC machine.
  \item \textbf{World model (ours)}: schema-adaptive, command-conditioned JEPA pretraining on the source data.
  \item \textbf{Compact representation of the dynamics}: a latent state that summarizes the machine's physical evolution.
  \item \textbf{Transfer to the target machine}: zero-shot use of the source-trained representation under partial sensor overlap.
  \item \textbf{Adaptation with a few target samples}: few-shot or online updates using limited data from the new machine.
  \item \textbf{World model}: the adapted latent dynamics used as a simulator of the target machine.
  \item \textbf{Simulation of candidate commands}: rolling out alternative command sequences in the world model.
  \item \textbf{Reinforcement learning}: policy optimization inside the learned simulator.
  \item \textbf{New control policy}: the resulting candidate policy for the target machine.
  \item \textbf{Doubly robust evaluation on historical data}: off-policy evaluation of the candidate policy with historical logs before any physical test.
  \item \textbf{Decision policy}: the evaluated policy that is retained for deployment or decision support.
\end{enumerate}
The present V1 paper covers steps 1--5 (steps 4 and 5 as zero-shot and pre-lock few-shot evidence). Figure~\ref{fig:v2} shows how V2 extends this backbone.

\section{Locked Configuration}\label{app:config}
\small
Table~\ref{tab:locked-config} lists the hyperparameters of the locked candidate (``schema + command recovery''; M03 in the code repository), the configuration frozen with SHA-256 hashes before the single confirmatory pass on the target machine.
\begin{table*}[t]\centering\caption{Locked configuration of the world model.}\label{tab:locked-config}
\footnotesize\setlength{\tabcolsep}{6pt}\renewcommand{\arraystretch}{1.0}
\begin{tabular}{@{}lll@{}}
\toprule
\rowcolor{cvprblue!10}
Group & Hyperparameter & Value \\
\midrule
\multirow{8}{*}{Architecture}
 & Latent dimension $d$ & 256 \\
 & Attention heads & 8 \\
 & Encoder layers (temporal transformer) & 6 \\
 & Predictor layers & 4 \\
 & Feed-forward width & 1024 \\
 & Dropout & 0.1 \\
 & Command injection & token \\
 & Transformer normalization & pre-norm \\
\midrule
\multirow{3}{*}{Windows}
 & Context length $K$ & 32 \\
 & Horizons $\mathcal{H}$ & $\{1,2,4,8,16\}$ \\
 & Normalization & source z-score (train split only) \\
\midrule
\multirow{3}{*}{Masking}
 & Mode & channel \\
 & Mask ratio \texttt{masking.ratio} $\rho$ & 0.35 (inactive in channel mode) \\
 & Channel drop probability \texttt{channel\_drop\_prob} & 0.15 \\
\midrule
\multirow{11}{*}{Objectives}
 & Latent loss & smooth $L_1$ \\
 & Latent weight $\lambda_{\mathrm{lat}}$ & 1.0 \\
 & Latent LayerNorm & off \\
 & VICReg mode / placement & per-horizon / context + target \\
 & VICReg weight $\lambda_{\mathrm{vic}}$ & 0.05 \\
 & VICReg coefficients (sim/var/cov) & 25 / 25 / 1 (invariance coefficient unused)  \\
 & Target aggregation & per-horizon \\
 & Schema-consistency weight $\lambda_{\mathrm{sch}}$ & 0.10 \\
 & Schema-view keep probability $\kappa$ & 0.65 (set in the trainer, not in the YAML) \\
 & Command-recovery weight $\lambda_{\mathrm{act}}$ & 0.05 \\
 & Physical weight (fine-tuning stage) & 1.0 \\
\midrule
\multirow{8}{*}{Optimization}
 & EMA momentum $\tau$ & 0.996 \\
 & Epochs (per stage) & 100 \\
 & Batch size & 128 \\
 & Learning rate & $3\times10^{-4}$ \\
 & Weight decay & 0.01 \\
 & Gradient-norm clip & 1.0 \\
 & Mixed precision & enabled \\
 & Early-stopping patience & 15 \\
\midrule
\multirow{3}{*}{Protocol}
 & Group split key & session \\
 & Fine-tuning head & fresh \\
 & Probabilistic head & Gaussian mean and log-variance \\
\midrule
\multirow{4}{*}{Instance norm.}
 & RevIN \texttt{model.revin.enabled} & off (locked); on in the variant of \hyperref[app:revin]{App.~H} \\
 & Affine parameters & none \\
 & Last-value subtraction & none \\
 & $\epsilon$ & $10^{-5}$ \\
\bottomrule
\end{tabular}
\end{table*}
\noindent Values from \texttt{configs/search\_v2/M03.yaml} in the code repository; the RevIN variant is \texttt{configs/revin/M03\_revin.yaml}, identical except for the flag.

\section{Paired RevIN Ablation}\label{app:revin}
\small
This appendix reports the post-lock ablation of Sec.~\ref{sec:revin-results} in full. Both arms were trained with \texttt{scripts/68\_revin\_ablation.py}, which runs the two-stage pipeline of \texttt{scripts/41} (self-supervised pretraining, fresh-head fine-tuning, common-space validation bundle) once per seed and arm: the control arm is \texttt{configs/search\_v2/M03.yaml} verbatim, the variant arm is the same file with \texttt{model.revin.enabled} set (\hyperref[app:config]{App.~G}). The three control re-runs reproduce the source-validation metrics of the locked model's seeds 0--2 exactly. Six training runs (3 seeds $\times$ 2 arms) of about 72 minutes each were executed sequentially on the GB10. Eighteen regression tests cover the transform (equivariance of the de-normalized mean under affine changes of the input, insensitivity of the statistics to masked entries, round-trip identity, identity on unseen channels, bitwise equality of the disabled path with the V1 code, and loading of V1 checkpoints).

\paragraph{Source validation.} Table~\ref{tab:revin-ds01} lists the nine validation components of the composite score in the common evaluation space, with the paired difference, its bootstrap 95\% interval over the three paired seeds, and the per-seed win count. The anchor row is the trivial-predictor RMSE of the evaluation space and is identical on both arms, which confirms that de-normalization returns to the same metric space. With $n=3$ pairs the exact sign-flip test cannot go below $p=0.125$; the intervals and win counts are the evidence. The latent-health gate flags the three RevIN runs as \texttt{ssl\_latent\_collapse} on its raw-scale sub-criterion only (Table~\ref{tab:revin-gate}); the scale-free sub-criterion is better with RevIN, latent health after fine-tuning is identical, and the gate was not modified.

\begin{table*}[t]\centering
\caption{Paired RevIN ablation on source validation (three seeds, mean $\pm$ SD; common evaluation space; lower is better except command sensitivity and effective rank). $\Delta$ is RevIN minus control; the interval is a bootstrap 95\% interval of the paired difference.}
\label{tab:revin-ds01}
\setlength{\tabcolsep}{5pt}
\begin{tabular}{@{}lccccc@{}}
\toprule
\rowcolor{cvprblue!10}
Component & Locked (control) & Locked + RevIN & $\Delta$ & 95\% interval & RevIN wins \\
\midrule
Clean RMSE & $0.8190\pm0.0103$ & $0.7663\pm0.0012$ & $-0.0527$ & $[-0.0640,\,-0.0457]$ & 3/3 \\
Schema-drop RMSE & $0.9180\pm0.0047$ & $0.8867\pm0.0034$ & $-0.0313$ & $[-0.0357,\,-0.0248]$ & 3/3 \\
Masked RMSE & $0.8272\pm0.0096$ & $0.7862\pm0.0023$ & $-0.0409$ & $[-0.0508,\,-0.0335]$ & 3/3 \\
Long-horizon RMSE & $0.8683\pm0.0073$ & $0.8487\pm0.0039$ & $-0.0196$ & $[-0.0234,\,-0.0167]$ & 3/3 \\
Calibration penalty $|\mathrm{cov}_{90}-0.9|$ & $0.1242\pm0.0270$ & $0.0349\pm0.0022$ & $-0.0892$ & $[-0.1201,\,-0.0691]$ & 3/3 \\
SSL validation loss & $1.1874\pm0.0189$ & $1.0640\pm0.0214$ & $-0.1234$ & $[-0.1660,\,-0.0952]$ & 3/3 \\
Cmd. sensitivity (shuffled / true) & $1.0639\pm0.0243$ & $1.1790\pm0.0013$ & $+0.1151$ & $[+0.1004,\,+0.1426]$ & 3/3 \\
Effective-rank fraction & $0.4266\pm0.0717$ & $0.4626\pm0.0062$ & $+0.0360$ & $[-0.0028,\,+0.1118]$ & 1/3 \\
Evaluation-space anchor RMSE & $0.9575\pm0.0000$ & $0.9575\pm0.0000$ & $0$ & $[0,\,0]$ & --- \\
\midrule
NLL (model space) & 3.02 & 1.49 & & & \\
Coverage at nominal 90\% & 0.776 & 0.865 & & & \\
\bottomrule
\end{tabular}
\end{table*}

\begin{table*}[t]
\begin{minipage}[t]{0.48\textwidth}\centering
\caption{Latent-health gate on the predicted latents (\texttt{trainers.latent\_health}). Thresholds: minimum standard deviation $<0.05$ (raw latent units, input-scale dependent) or effective-rank fraction $<0.10$ (scale-free). SSL: after pretraining; FT: after fine-tuning.}
\label{tab:revin-gate}
\footnotesize\setlength{\tabcolsep}{3pt}
\begin{tabular}{@{}llcccc@{}}
\toprule
\rowcolor{cvprblue!10}
Arm & Seed & Std$_{\min}$ SSL & Rank SSL & Std$_{\min}$ FT & Rank FT \\
\midrule
Locked & 0 & 0.061 & 0.223 & 0.391 & 0.468 \\
Locked & 1 & 0.060 & 0.227 & 0.327 & 0.344 \\
Locked & 2 & 0.059 & 0.231 & 0.365 & 0.468 \\
Locked + RevIN & 0 & 0.040 & 0.488 & 0.280 & 0.467 \\
Locked + RevIN & 1 & 0.041 & 0.470 & 0.291 & 0.456 \\
Locked + RevIN & 2 & 0.041 & 0.446 & 0.294 & 0.465 \\
\bottomrule
\end{tabular}
\end{minipage}\hfill
\begin{minipage}[t]{0.48\textwidth}\centering
\caption{Second, declared target read: every run of the paired ablation on the 2,457 JOANNEUM windows (10 shared channels). Persistence and the confirmatory pass of the locked model are given for reference.}
\label{tab:revin-ds03}
\footnotesize\setlength{\tabcolsep}{3pt}
\begin{tabular}{@{}llcccc@{}}
\toprule
\rowcolor{cvprblue!10}
Arm & Seed & RMSE & MAE & $R^2$ & NLL \\
\midrule
Persistence & --- & 0.654 & 0.316 & $-0.420$ & --- \\
Locked, confirmatory pass & 0 & 0.546 & --- & $+0.012$ & 0.52 \\
\midrule
Locked & 0 & 0.5456 & 0.3524 & $+0.0118$ & 0.516 \\
Locked & 1 & 0.5716 & 0.3682 & $-0.0847$ & $-0.002$ \\
Locked & 2 & 0.5470 & 0.3496 & $+0.0067$ & 2.147 \\
Locked + RevIN & 0 & 0.4994 & 0.2796 & $+0.1720$ & 22.48 \\
Locked + RevIN & 1 & 0.4924 & 0.2722 & $+0.1950$ & 15.07 \\
Locked + RevIN & 2 & 0.4928 & 0.2753 & $+0.1937$ & 24.35 \\
\bottomrule
\end{tabular}
\end{minipage}
\end{table*}

\paragraph{Declared second target read.} The lock consumed the single declared target pass. The second opening of the JOANNEUM windows was decided before the ablation ran and conditioned on the variant winning the four RMSE components of Table~\ref{tab:revin-ds01} on source validation; that verdict was written to \texttt{summary.json} before the target stage started, and the target stage then scored all six runs with the protocol of \texttt{scripts/45} (model rebuilt from the run configuration, source-train normalizers, 2,457 windows, 10 shared channels) and no selection on the target. Table~\ref{tab:revin-ds03} lists every run. Aggregates: control RMSE $0.555\pm0.015$, MAE $0.357$, $R^2=-0.02$, NLL $0.89$; RevIN RMSE $0.495\pm0.004$, MAE $0.276$, $R^2=+0.19$, NLL $20.6$; paired RMSE differences $-0.046$, $-0.079$, $-0.054$. Coverage at the nominal 90\% level is 0.953 for the control and 0.874 for RevIN. The per-channel diagnosis of the NLL increase (seed 0: 11.5\% of predictions at the log-variance floor; spindle torque 89 nats, Z power 51, spindle current 40, spindle power 28) and its mechanism are given in Sec.~\ref{sec:revin-results}.

\end{document}